%% file: main.tex
\documentclass[Afour,sageh,times]{sagej}
\input{settings.tex}
\begin{document}
\runninghead{Pushp \textit{et al.}}
\title{Navigating the Wild: Pareto-Optimal Visual Decision-Making in Image Space
}
\author{}
\author{Durgakant Pushp\affilnum{1},
    Weizhe Chen\affilnum{1},
    Zheng Chen\affilnum{1},
    Chaomin Luo\affilnum{2},
    Jason M. Gregory\affilnum{3}, and
    Lantao Liu\affilnum{1}%
}%
\affiliation{%
    \affilnum{1} D. Pushp, W. Chen, Z. Chen and L. Liu are with Luddy School of Informatics, Computing, and Engineering, Indiana University, Bloomington, IN 47408, USA. Email: {\tt\small dpushp@iu.edu}\\
    \affilnum{2} C. Luo is with the Department of Electrical and Computer Engineering, Mississippi State University, MS 39762, USA.\\
    \affilnum{3} J. Gregory is with DEVCOM Army Research Laboratory, USA.%
}%
\corrauth{Lantao Liu}
\email{lantao@iu.edu}

\begin{abstract}
Navigating complex real-world environments requires understanding the semantic context and effectively making decisions. Existing solutions leave room for improvements: traditional reactive approaches that do not maintain a map often struggle in complex environments, map-dependent methods demand significant effort in mapping processes, and learning-based methods rely on large training datasets and face the difficulty of generalization. To address these challenges, we propose a novel visual semantic navigation framework that combines data-driven semantic understanding, Pareto-optimal decision-making, and image-space planning. Our approach uses a local environmental representation called \textit{navigability image}, which allows the robot to assess immediate traversability without relying on \textit{apriori} mapping or navigation data. Building on this, we introduce \textit{Pareto-Optimal Visual Navigation (POVNav)}, a decision-making framework in the image space that identifies appropriate sub-goals, constructs collision-free paths, and generates control commands using visual servoing. This framework also supports selective navigation behaviors, such as avoiding traversable yet slippery grasslands to prevent getting stuck, by dynamically adjusting the navigability criteria within the local representation. POVNav is lightweight, operating solely with a monocular camera and without requiring map storage or training data collection, making it highly versatile for different robotic platforms and environments. Extensive year-round real-world experiments validated its efficacy in both structured indoor environments and unstructured outdoor settings, including dense forest trails and snow-covered roads. Field experiments using various image segmentation techniques demonstrated its robustness and adaptability across a wide range of conditions. Additionally, we demonstrate that POVNav sucessfully guides a robot through narrow pipes in a culvert inspection task. Overall, we showcase the utility of POVNav in real-world scenarios, highlighting its flexibility and computational efficiency for autonomous robots in complex environments.
\end{abstract}

\keywords{Navigation in the Wild, Visual Semantic Navigation, Pareto-Optimal Decision-Making}

\maketitle
\input{1_introduction}
\input{2_literature}
\input{3_background}
\input{4_methodology}
\input{5_experiments}

\input{6_conclusion}
\bibliographystyle{SageH}
\interlinepenalty=10000
\bibliography{references.bib}
\end{document}

%% file: settings.tex
\usepackage{moreverb,url}
\usepackage[colorlinks,bookmarksopen,bookmarksnumbered,citecolor=red,urlcolor=red]{hyperref}
\newcommand\BibTeX{{\rmfamily B\kern-.05em \textsc{i\kern-.025em b}\kern-.08em
T\kern-.1667em\lower.7ex\hbox{E}\kern-.125emX}}
\def\volumeyear{2023}
\usepackage{bm}
\usepackage{amssymb}
\usepackage{amsthm}
\usepackage{mathtools}
\usepackage{nicefrac}  

\usepackage[noend]{algpseudocode}
\usepackage{algorithm}
\usepackage{setspace}

\usepackage{booktabs}
\usepackage{multicol}
\usepackage{multirow}
\usepackage{tabularx}
\usepackage{etoolbox}
\newrobustcmd{\B}{\bfseries}  
\usepackage{siunitx}  
\usepackage{diagbox}
\usepackage{pifont}  
\usepackage{todo}  
\usepackage[table]{xcolor}
\definecolor{red}{RGB}{200,50,50}
\definecolor{green}{RGB}{50,200,50}
\definecolor{blue}{RGB}{50,50,200}
\definecolor{orange}{RGB}{243,147,82}
\definecolor{purple}{RGB}{150,100,250}
\definecolor{yellow}{RGB}{231,198,100}
\definecolor{gray}{RGB}{98,95,75}
\definecolor{white}{RGB}{255,255,255}
\usepackage{lipsum}
\usepackage{times}

\graphicspath{{./Figures/}}

\makeatletter
\newcommand*{\T}{{\mathpalette\@transpose{}} }
\newcommand*{\@transpose}[2]{\raisebox{\depth}{$\m@th#1\intercal$}}
\makeatother
\DeclarePairedDelimiterX{\norm}[1]{\lVert}{\rVert_{2}}{#1}

\DeclareMathOperator*{\argmin}{arg\,min}

\makeatletter
\newcommand\thefontsize{The current font size is: \f@size pt}
\makeatother

\usepackage{cleveref}  
\theoremstyle{definition}
\newtheorem{definition}{Definition}
\newtheorem{theorem}{Theorem}

\newtheorem{assumption}{Assumption}

\newtheorem{remark}{Remark}

\crefname{proposition}{Proposition}{Propositions}

\newtheorem{lemma}[theorem]{Lemma}
\usepackage{balance}  
\apptocmd{\sloppy}{\hbadness 10000\relax}{}{}

\usepackage{graphicx}
\usepackage{subfigure}
\usepackage{wrapfig}  
\usepackage{tikz}
\usetikzlibrary{positioning,shapes,bayesnet,calc,backgrounds}

\makeatletter
\newif\if@showgrid@grid
\newif\if@showgrid@left
\newif\if@showgrid@right
\newif\if@showgrid@below
\newif\if@showgrid@above
\tikzset{%
  every show grid/.style={},
  show grid/.style={execute at end picture={\@showgrid{grid=true,#1}}},%
  show grid/.default={true},
  show grid/.cd,
  labels/.style={font={\sffamily\small},help lines},
  xlabels/.style={},
  ylabels/.style={},
  keep bb/.code={\useasboundingbox (current bounding box.south west) rectangle (current bounding box.north west);},
  true/.style={left,below},
  false/.style={left=false,right=false,above=false,below=false,grid=false},
  none/.style={left=false,right=false,above=false,below=false},
  all/.style={left=true,right=true,above=true,below=true},
  grid/.is if=@showgrid@grid,
  left/.is if=@showgrid@left,
  right/.is if=@showgrid@right,
  below/.is if=@showgrid@below,
  above/.is if=@showgrid@above,
  false,
}
\def\@showgrid#1{%
  \begin{scope}[every show grid,show grid/.cd,#1]
    \if@showgrid@grid
      \begin{pgfonlayer}{background}
        \draw [help lines]
        (current bounding box.south west) grid
        (current bounding box.north east);
        \pgfpointxy{1}{1}%
        \edef\xs{\the\pgf@x}%
        \edef\ys{\the\pgf@y}%
        \pgfpointanchor{current bounding box}{south west}
        \edef\xa{\the\pgf@x}%
        \edef\ya{\the\pgf@y}%
        \pgfpointanchor{current bounding box}{north east}
        \edef\xb{\the\pgf@x}%
        \edef\yb{\the\pgf@y}%
        \pgfmathtruncatemacro\xbeg{ceil(\xa/\xs)}
        \pgfmathtruncatemacro\xend{floor(\xb/\xs)}
        \if@showgrid@below
          \foreach \X in {\xbeg,...,\xend} {
            \node [below,show grid/labels,show grid/xlabels] at (\X,\ya) {\X};
          }
        \fi
        \if@showgrid@above
          \foreach \X in {\xbeg,...,\xend} {
            \node [above,show grid/labels,show grid/xlabels] at (\X,\yb) {\X};
          }
        \fi
        \pgfmathtruncatemacro\ybeg{ceil(\ya/\ys)}
        \pgfmathtruncatemacro\yend{floor(\yb/\ys)}
        \if@showgrid@left
          \foreach \Y in {\ybeg,...,\yend} {
            \node [left,show grid/labels,show grid/ylabels] at (\xa,\Y) {\Y};
          }
        \fi
        \if@showgrid@right
          \foreach \Y in {\ybeg,...,\yend} {
            \node [right,show grid/labels,show grid/ylabels] at (\xb,\Y) {\Y};
          }
        \fi
      \end{pgfonlayer}
      \fi
    \end{scope}
  }
  \makeatother
  \tikzset{every show grid/.style={show grid/keep bb}%
  }%

\usepackage[most]{tcolorbox}
\usepackage{xcolor}

\newtcolorbox{modifyblock}{
  colback=orange!5!white,    
  colframe=orange,  
  boxrule=0.5pt,
  arc=1pt,
  left=0pt, right=0pt, top=0pt, bottom=0pt
}

\newtcolorbox{addblock}{
  colback=blue!5!white,    
  colframe=blue,  
  boxrule=0.5pt,
  arc=1pt,
  left=0pt, right=0pt, top=0pt, bottom=0pt
}

%% file: 1_introduction.tex
\section{Introduction}
\begin{figure*}
    \begin{center}
        \includegraphics[width=\textwidth,height=0.45\textwidth]{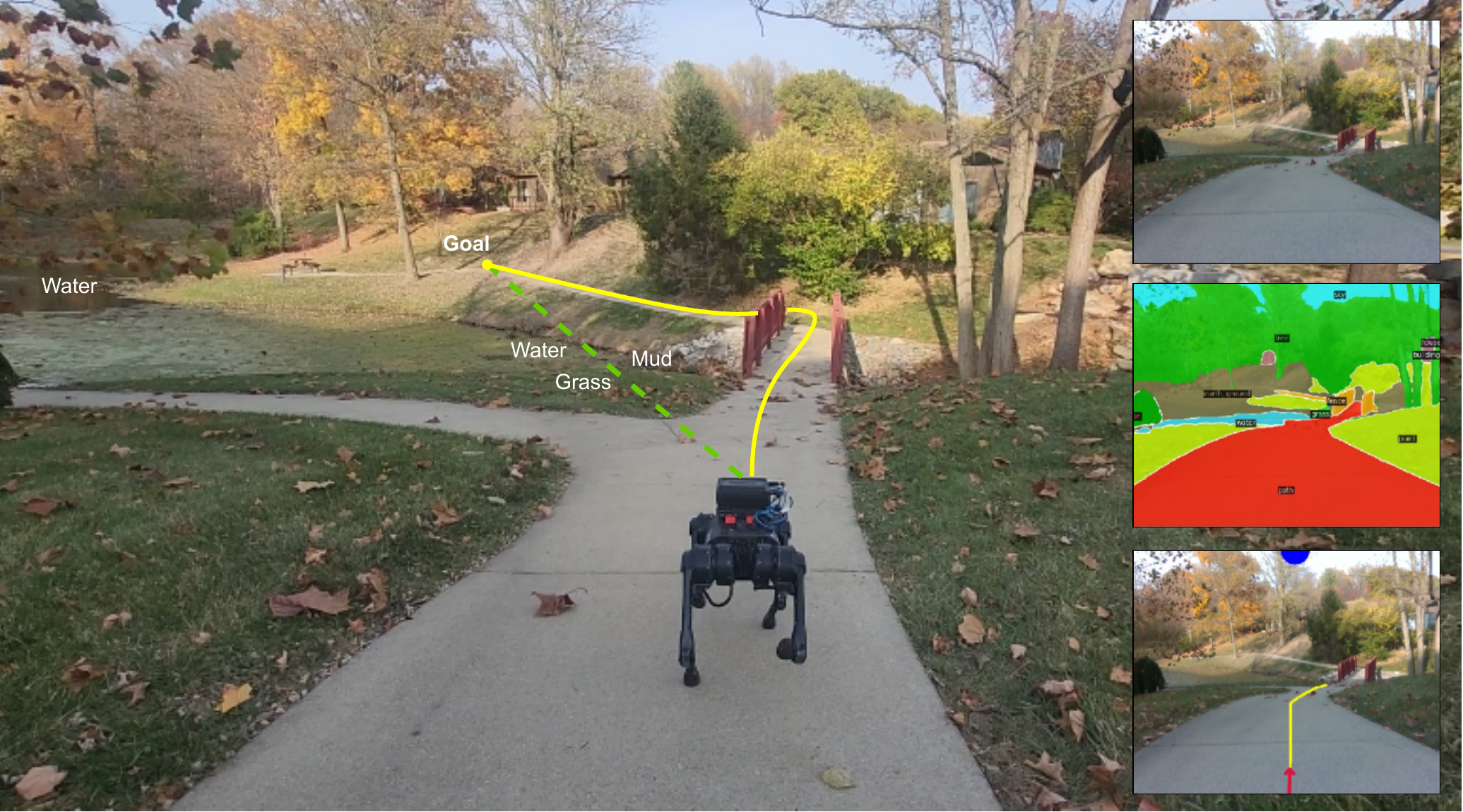}
    \end{center}
    \caption{A motivating example of visual semantic navigation in the wild. Visual semantic navigation enables the robot to interpret the semantic meaning of environmental elements and adapt its path to diverse terrain conditions. The green path represents the shortest route, passing through mud and water, while the yellow path avoids these non-traversable areas. On the right panels, from top to bottom, appear the observed image, segmented image, and a planning image illustrating a semantically-aware, safe path.}
    \label{fig:motivation}
\end{figure*}

Humans possess a remarkable ability to navigate complex environments by intuitively interpreting visual scenes at a semantic level -- effortlessly distinguishing between walkable paths, obstacles, and hazardous areas while adapting to diverse terrain conditions~\citep{dwivedi2024visual}. This natural ability to understand both the semantic meaning and traversability of environmental elements has inspired the development of visual semantic navigation systems for autonomous robots. Through semantic segmentation of the environment, robots can identify traversable spaces and obstacles, moving closer to achieving human-like navigation capabilities in challenging real-world applications. A motivating scenario is shown in~\Cref{fig:motivation}.

Visual semantic navigation is especially crucial in field robotics applications. For instance, in disaster response scenarios such as radiation accidents or search-and-rescue operations after devastating earthquakes, robots must quickly differentiate between safe terrain and hazardous areas, including fires and unstable structures~\citep{wellhausen2020safe}. Agricultural robots need to differentiate between crop rows and traversable paths covered with leaves and weeds, enabling efficient farmland navigation while preventing crop damage~\citep{sivakumar2021learned}. In marine environments, underwater robots must identify safe passages through coral reefs while maintaining stable navigation -- simultaneously focusing on scientifically relevant features and avoiding disturbance to sensitive ecosystems~\citep{koreitem2020one}.

In recent years, visual semantic navigation has witnessed significant advancements, with a wide range of approaches proposed to address the challenges of autonomous navigation in complex environments.
These approaches can be broadly categorized into two main ideas: modular and end-to-end approaches, each with distinct strengths and limitations. Modular methods are the classic or symbolic approaches that decompose the navigation task into separate modules, often including semantic Simultaneous Localization and Mapping~(SLAM), planning, and control. This modularity enhances system interpretability and allows for targeted improvements within each component. However, maintaining an accurate map in modular approaches can be expensive in terms of computation and storage. 
On the other hand, end-to-end methods bypass explicit mapping by using deep neural networks to directly transform raw sensor data into control commands, enabling fast and reactive decision-making. Although end-to-end methods reduce mapping overhead, they often lack interpretability and struggle to generalize well in unseen environments.

Deploying these approaches in the wild presents additional challenges. Outdoor environments are characterized by high variability and complexity, such as fluctuating lighting, changing weather, and diverse terrain, which can significantly impact system performance. Modular systems may find it costly to continuously update (semantic) maps which change constantly in dynamic environments, while end-to-end systems may fail to adapt effectively to unseen scenarios or sensor noise. These challenges highlight the need for a hybrid approach that enjoys the \emph{interpretability} and \emph{modularity} of modular systems and the \emph{adaptability} and \emph{efficiency} of end-to-end systems, combining the best of both worlds to achieve robust and efficient navigation in the wild.

To this end, we propose a novel framework for visual semantic navigation that leverages the strengths of both modular and end-to-end approaches.
To maintain interpretability and modularity, the proposed framework still comprises separate modules for perception, planning, and control, allowing for easy integration of new algorithms in each component, e.g., advanced segmentation models~\citep{kirillov2023segment} or reasoning traversability by Vision Language Models~(VLMs)~\citep{song2024tgs}. Instead of depending on explicit semantic mapping, we adopt an approach that directly utilizes image observations to guide navigation decisions, aligning with end-to-end methodologies.

Without a semantic map, traditional planning algorithms cannot be directly applied, since they require a global bird's-eye view representation of the environment. The challenge then becomes how to plan and control a robot using only ego-centric image. 
A straightforward solution is to segment the image into navigable and non-navigable regions, plan a path through the navigable regions from the robot's current position to the projected goal in the image, and control the robot to follow this path using visual servoing. However, this approach has two challenges:
\begin{itemize}
    \item \textbf{Segmentation Noise}: real-world segmentation images are typically noisy and incomplete, leading to suboptimal paths and control actions.
    \item \textbf{Goal Occlusion}: When the goal is occluded (e.g., behind a tree), there does not exist a collision-free path connecting the robot to the goal in the image space.
\end{itemize}

To address the first challenge, we observe that segmentation accuracy exhibits an inverse relationship with depth, i.e,  objects in close proximity yield more reliable segmentation results compared to distant scenes. Since distant objects typically have minimal impact on immediate navigation decisions, we introduce the concept of \emph{visual horizon} that marks the boundary of reliably navigable regions in the robot's vicinity. By constraining our analysis to the space within this visual horizon, we mitigate the effects of segmentation noise and incompleteness.

The second challenge can be addressed by selecting a subgoal in the image. Effective subgoal selection must balance multiple potentially competing criteria: (1) the subgoal must be reachable via an obstacle-free path; (2) reduce the distance to the final goal; and (3) allow sufficient deviation to navigate around obstacles when necessary. The proposed framework employs Pareto-optimality to analyze the trade-offs among these objectives, hence the name {\em Pareto-Optimal Visual Navigation~(POVNav)}. 

Intuitively, POVNav first generates a navigability image from the semantic segmented image, focusing on the space within the visual horizon. After projecting the goal onto the navigability image, POVNav selects a subgoal on the visual horizon that balances goal-reaching and obstacle avoidance. A simple and efficient planning algorithm then generates a safe visual path from the robot's current position to the subgoal, which is then followed using visual servoing. The advantages of POVNav are multi-fold:
\begin{itemize}
    \item \textbf{Modularity}: The proposed framework maintains the interpretability and modularity of traditional modular systems, allowing for easy integration of various advanced algorithms in each component.
    \item \textbf{Efficiency}: By directly processing image observations, POVNav avoids the overhead of maintaining an explicit map, enabling fast and reactive decision-making.
    \item \textbf{Robustness}: The concept of the visual horizon mitigates the effects of segmentation noise and incompleteness, enhancing the robustness of navigation decisions.
    \item \textbf{Adaptability}: By selecting subgoals on the visual horizon, POVNav can effectively navigate in dynamic environments with occluded goals, adapting to different scenarios.
\end{itemize}

This paper extends our prior conference publication \citep{pushp2023povnav} by substantially enriching the framework, analysis, and evaluation of POVNav. The main contributions of this journal version are:
\begin{itemize}
\item A comprehensive literature review situates POVNav within the broader context of vision-based navigation, highlighting its unique formulation that combines semantic segmentation, image-space planning, and Pareto-optimal decision-making.
\item New theoretical analysis is introduced, including proofs of weak Pareto-optimality, control stability, and computational complexity of the overall framework.
\item Additional experimental results are provided through extensive simulations and long-duration field trials across diverse indoor and outdoor environments using multiple robot platforms.
\item Systematic ablation and robustness studies isolate the impact of key components and validate the method's generalizability under varying environmental and perceptual conditions.
\item Expanded discussion of implementation details offers practical insights into segmentation model integration, parameter tuning, and real-time deployment strategies for resource-constrained systems.
\end{itemize}

Experimental results from comparing the proposed framework against representative baseline methods  reveal that, the proposed approach achieves a significantly higher success rate across diverse environments, ranging from sparsely populated areas to densely cluttered obstacle configurations. Moreover, POVNav consistently identifies and follows shorter paths. We also present real-world field trials to show the robustness and adaptability of the proposed framework, highlighting its potential to navigate through various challenging scenarios. Last but not least, we release the code to facilitate future research. 

%% file: 2_literature.tex
\section{Related Work}

Our work aligns with robust visual semantic navigation approaches, which generally fall into two main categories: non-learning-based methods, such as optical flow, appearance-based tracking, and object feature tracking; and learning-based solutions. In the following sections, we review key advancements in each of these areas, examining their contributions and limitations within the context of autonomous navigation. Additionally, as our proposed framework relies heavily on segmentation techniques to interpret and navigate environments, we discuss various semantic segmentation methods, highlighting how their differing outputs can impact navigational performance.

\subsection{Feature-Based Visual Navigation}
Visual features are crucial in enabling robots to interpret and interact with their surroundings. Optical flow, as a widely-used visual feature, plays a significant role in obstacle avoidance and visual navigation. A comprehensive review of optical flow methods is provided by \cite{hongche1998optical}. \cite{giachetti1998optical} introduced a method for achieving dense and reliable optical flow from images captured by a car-mounted camera in outdoor environments. By employing correlation-based techniques and shock compensation, they estimated egomotion, including speed and angular velocity, through analysis of flow patterns on flat roads. This optical flow approach also facilitated coarse segmentation, allowing for the differentiation of moving objects and enhancing the localization of motion boundaries. These findings highlight the potential of optical flow in vision systems for vehicle navigation and driver assistance applications.

\cite{kai2004optical} combines correlation-based and differential techniques to achieve faster real-time estimation. Optical flow has since been widely applied in moving object detection~\citep{talukder2004optical, braillon2006optical} and for mobile robot navigation within constrained environments~\citep{mccarthy2004optical}. \cite{ohnishi2006optical} applied optical flow to detect dominant planes — a critical element in path planning. Furthermore, optical flow algorithms have demonstrated significant potential for real-time UAV navigation across a range of tasks~\citep{yamada2003opticaluav, rondon2009opticaluav, conroy2009opticalimplementation, mammarella2008opticalcomparison, janschek2006opticalperformance}. Specifically, \cite{zingg2010opticalmav} developed a wall collision avoidance technique utilizing a depth map derived from optical flow on images captured by onboard cameras. 

The growing popularity of optical flow has spurred developments in both hardware and open-source solutions, such as the optical flow sensor designed by \cite{honegger2013opticalmav}, which integrates a machine vision CMOS image sensor and has been demonstrated in-flight on a micro air vehicle. Over the past decade, research has increasingly focused on refining optical flow feature estimation through diverse methods~\citep{zhai2021opticalestimation, hui2018opticalestimation, ilg2017opticalestimation, ilg2018opticalestimation, ren2017opticalestimation, wulff2015opticalestimation, bailer2015opticalestimation, ranjan2017opticalestimation}, leading to notable improvements in visual navigation. However, relatively few studies have examined the integration of optical flow into motion planning and control. Recent works that address this gap include \cite{ponce2018optical, károly2019optical, meronen2020optical, argus2020optical, de2021optical, boretti2022ttt}, highlighting the emerging interest in leveraging optical flow for comprehensive navigation and control in robotics. In general, optical flow-based methods are effective for obstacle avoidance as long as they can detect sufficient features. However, they lack the ability to perform goal-oriented navigation and adaptive behavior.

While optical flow has remained a popular feature for visual navigation over the past decades, previous research has also explored alternative approaches, including stored image templates~\citep{bista2016appearance, hong1992imagetemplate, van2002visualtemplate, chang2010mobiletemplate, mielle2016usingtemplate, courbon2008indoortemplate, menegatti2004imagetemplate, franz2000biomimetictemplate, srinivasan1999robottemplate, jayender2008autonomoustemplate}, information-theoretic approach \citep{dame2011new, vakil2015information, wu2021reinforcementinformation, carlone2018attentioninformation, soundararajan2013surveyinformation} and feature tracking-based navigation methods~\citep{anand2019gaussian,srivastava2021estimation,pushp2022uav, jia2009visiontracking, wang2021visualtracking, zhao2013objecttracking, mateus2005robottracking, zhang2021visualtracking, liu2023efficienttracking} to control robot motion. 
An image segmentation approach based on sub-region growing to determine paths is proposed by \cite{zhang2008autonomous}, where control parameters are derived from path boundary information. Although these methods provide solutions for obstacle avoidance, they fall short in facilitating goal-directed navigation. Achieving visual, goal-oriented navigation poses unique challenges, as existing techniques struggle to relate 3D goal locations to the observable image features, particularly when the goal lies outside the camera’s field of view. 
A notable prior work that explores a related direction is~\cite{otte2009path}, which proposed planning directly in image space using disparity-based representations to generate paths without relying on global maps. While this work demonstrated the feasibility of image-space planning and local obstacle avoidance, it did not account for non-navigability arising from semantic understanding and focused solely on geometric cues derived from depth information.
In contrast, our proposed framework incorporates semantic information to reason about navigability, and identifies an optimal sub-goal within the current observation, effectively bridging the gap between image-space features and goal-directed behavior. Unlike previous methods, our approach provides a unified framework for navigation in unknown environments, relying solely on a monocular camera and a directional cue toward the goal.

\subsection{Learning-Based Visual Navigation}

A large family of current frameworks is data-driven or based on machine learning \citep{shen2019situational,bansal2020combining}. For example, imitation learning-based approaches have been broadly explored to train a navigation policy that enables a robot to mimic human behaviors or navigate closely to certain waypoints without a prior map~\citep{manderson2020vision,hirose2021probabilistic,pan2020zeroimitation, wu2020towardsimitation, yan2022maplessimitation, karnan2022voilaimitation}. 
A large amount of work on visual navigation can also be found in the computer vision community~\citep{chaplot2020neural, lei2024instance, kwon2023renderable, wasserman2023last, kim2023topological, hahn2021no, kwon2021visual, li2020learning, chaplot2020neural, liu2022symmetry, chaplot2020semantic}. These approaches exploit and integrate deep learning methods to train navigation policies that perform remarkably well when provided sufficient training data, but can fail if no or very limited data is available. In contrast to the data-driven approaches, we introduce a computationally light but effective planning and control loop that decouples the representation from the behavior learning (like modular learning approaches) using segmentation-based visual representation. Our solution is more flexible, robust, and adaptable to a variety of navigation missions since it makes no assumption about the data it needs to learn the policy. 

\subsection{Visual Semantic Segmentation}
Semantic segmentation assigns a unique human-defined semantic label to each pixel in an image, providing detailed scene understanding essential for applications like navigation. Our proposed method, POVNav, leverages pre-trained segmentation models to parse navigability and uses image-space planning for action generation. The overall performance of POVNav is closely tied to the predictive accuracy of the employed segmentation models.

The advent of deep neural networks, particularly Fully Convolutional Networks (FCNs)~\citep{long2015fully}, significantly improved semantic segmentation performance by using convolutional neural networks to predict pixel-wise classifications. Subsequent works have enhanced FCNs through various techniques, such as multi-scale feature aggregation using different kernel sizes and dilation rates~\citep{chen2017deeplab, chen2017rethinking, yu2015multi}, multi-resolution input generation via image pyramids~\citep{zhao2017pyramid}, probabilistic graph-based smoothing~\citep{liu2017deep}, and encoder-decoder structures for feature compensation~\citep{ronneberger2015u}. 
Recently, Transformer-based models~\citep{vaswani2017attention, dosovitskiy2020image, ranftl2021vision} have gained prominence in semantic segmentation. Transformers utilize attention mechanisms to model long-range dependencies among pixels. Methods such as \cite{zheng2021rethinking} treat semantic segmentation as a sequence-to-sequence prediction task, encoding images as patch sequences. SegFormer~\citep{xie2021segformer} combines a hierarchical Transformer encoder with a lightweight MLP decoder to create a robust yet efficient segmentation framework. 
A novel architecture for image segmentation~\citep{cheng2022masked} incorporates masked attention, which extracts localized features by constraining cross-attention within predicted mask regions. These methods are particularly suited for applications like POVNav, as they allow online correction of navigable and non-navigable class definitions, addressing domain shifts effectively. However, field experiments revealed significant performance degradation when there is a substantial difference between the training and testing data, highlighting their limitations under extreme domain shifts.

To address domain adaptation challenges, transfer learning frameworks have been widely explored. Techniques such as adversarial training~\citep{ganin2016domain, hoffman2016fcns, hoffman2018cycada, chen2022cali, saito2018maximum, luo2019taking, wang2020classes, chen2023pseudo, khan2025afrda} align data distributions between source and target domains, while self-training methods~\citep{zou2018unsupervised, mei2020instance, hoyer2021daformer, zou2019confidence, xie2022towards, chen2019domain, hoyer2022hrda, hoyer2022daformer, chen2023ida} iteratively refine predictions on target domain data, demonstrating robustness in real-world scenarios.

For our proposed POVNav framework, we experimented with various semantic segmentation techniques, including Transformer-based and domain-adaptive methods, to evaluate their impact on navigation performance. Field experiments revealed that while state-of-the-art segmentation models generally enhance navigability predictions, they may fail under significant domain shifts. To mitigate this, POVNav dynamically adjusts navigable and non-navigable class definitions online, improving adaptability across diverse environments. Our findings highlight the critical role of segmentation accuracy in achieving reliable and efficient navigation.

%% file: 3_background.tex
\section{Background}

\begin{table}[tbp]
    \caption{\textbf{Notation System}.}\label{tab:notation_system}
  \centering
  \small
  \begin{tabular}{rrr}
    \toprule
    \textbf{Meaning}           & \textbf{Example}           & \textbf{Remark}            \\
    \midrule
    variable            & $x$                 & lower-case          \\
    constant            & $X$                 & upper-case          \\
    vector              & $\mathbf{x}$        & bold, lower-case    \\
    matrix              & $\mathbf{X}$        & bold, upper-case    \\
    set or space           & $\mathbb{R}$        & blackboard, upper-case\\
    function            & $\mathtt{f}(\cdot)$ & typewriter          \\
    functional          & $\mathtt{E}(\cdot)$ & typewriter, upper-case\\
    special PDF         & $\mathcal{N}$       & calligraphy, upper-case\\
    definition          & $\triangleq$        & \\
    transpose           & $\mathbf{m}^{\T}$   &  \\
    absolute value      & $|\cdot|$ & \\
    Euclidean norm      & $\norm{\cdot}$ &  \\
    \bottomrule
  \end{tabular}%
\end{table}

\begin{table}[tbp]
    \caption{\textbf{Mathematical Symbols}.}\label{tab:math_symbols}
  \centering
  \small
  \begin{tabular}{rr}
    \toprule
    \textbf{Symbol} & \textbf{Meaning}\\
    \midrule
    $\mathbb{E}$ & environment \\
    $t$ & time step \\
    $\mathbf{s}_{t}$ & robot's state\\
    $\mathbf{p}_{t}$ & robot's position\\
    $x_t$ & x-coordinate of the robot \\
    $y_t$ & y-coordinate of the robot\\
    $\theta_t$ & heading angle of the robot\\
    $\mathbf{I}_t$ & image captured by the robot\\
    $H$ & height of the image\\
    $W$ & width of the image\\
    $\mathbf{a}_t$ & robot's action\\
    $v_t$ & linear velocity of the robot\\
    $\omega_t$ & angular velocity of the robot\\
    $\mathbf{g}$ & goal position\\
    $x_g$ & x-coordinate of the goal\\
    $y_g$ & y-coordinate of the goal\\
    $\mathbb{O}_{t}$ & non-navigable areas\\
    $\pi$ & policy\\
    $\mathtt{f}(\cdot,\cdot)$ & transition function\\
    $\mathtt{d}(\cdot,\cdot)$ & distance function\\
    $D_\text{safe}$ & safety distance\\
    $V_\text{max}$ & maximum linear velocity\\
    $\Omega_\text{max}$ & maximum angular velocity\\
    $\Delta V_\text{max}$ & maximum linear acceleration\\
    $\Delta \Omega_\text{max}$ & maximum angular acceleration\\
    $T$ & time to reach the goal\\
    $\varepsilon$ & goal region radius\\
    $\mathbf{I}_{\text{seg}}$ & segmented image\\
    $\mathbf{I}_{\text{nav}}$ & navigability image\\
    $\mathbb{H}$ & visual horizon pixels\\
    $\mathbb{G}$ & subgoal search space\\
    $\mathbf{c}$ & cost vector\\
    $\mathtt{c}_\text{nav}$ & navigation cost\\
    $\mathtt{c}_\text{exp}$ & exploration cost\\
    $w$ & weight parameter\\
    $\boldsymbol{\xi}$ & safe visual path\\
    $\lambda$ & min distance to visual horizon\\
    $\phi$ & angular deviation from safe path\\
    \bottomrule
  \end{tabular}%
\end{table}

In this section, we introduce the assumptions, problem formulation, and notation system used in this work. \Cref{tab:notation_system} summarizes the notation system, and \Cref{tab:math_symbols} lists the mathematical symbols used in this paper.

\subsection{Assumptions}

In this work, we make some reasonable assumptions to simplify the problem and focus on the core challenges of visual navigation.

\begin{assumption}[Relative Position Estimation] \label{assumption:relative}
    The robot can estimate its relative direction to the goal. This can be achieved by using practical methods such as visual data interpretation, odometry, or a combination of visual inputs with inertial and/or GPS signals (if available) for improved accuracy.
\end{assumption}
\begin{assumption}[Camera Calibration]\label{assumption:camera}
    Without loss of generality, a monocular RGB camera is \emph{calibrated} and \emph{centered} at the robot's position.
\end{assumption}
\begin{assumption}[Velocity Control]
    The robot's motion can be controlled through velocity commands.
\end{assumption}
\begin{assumption}[Reasonable Obstacle Speeds]
    The dynamic obstacles move at speeds that allow for reactive navigation.
\end{assumption}

\begin{figure*}[t!]
    \begin{center}
        \includegraphics[width=\linewidth]{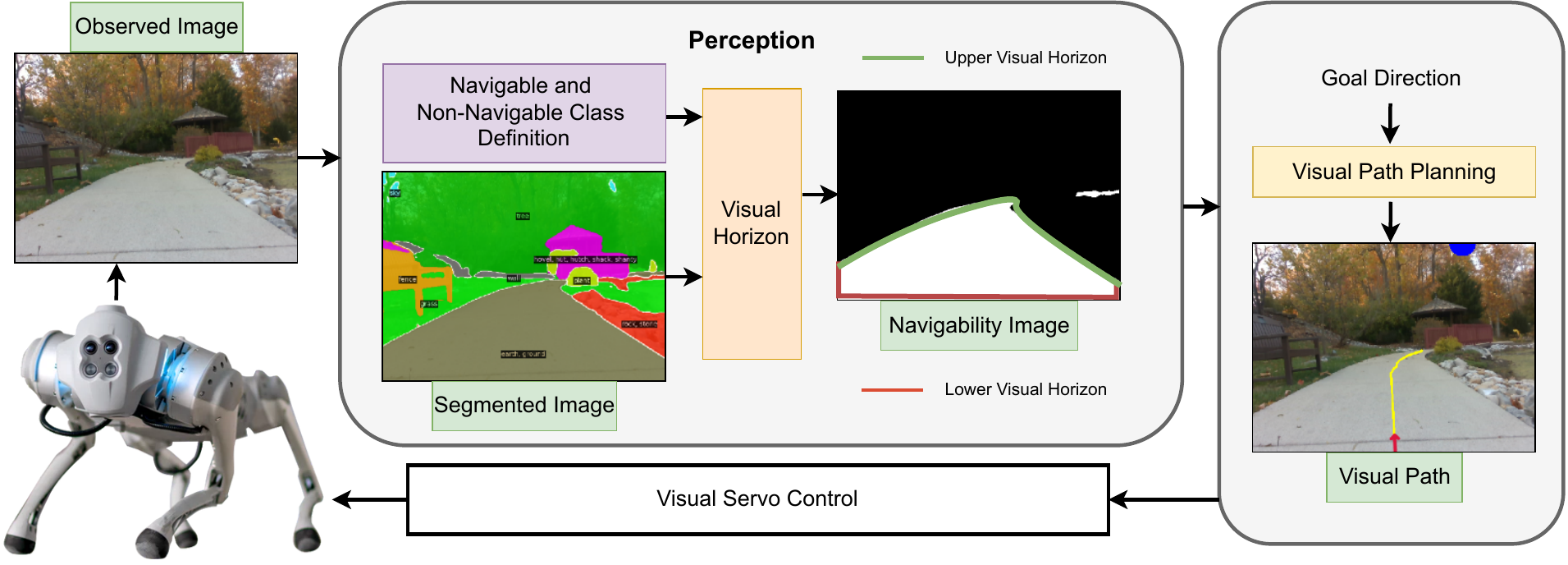}
    \end{center}
    \caption{System Diagram of the Proposed Framework. The system diagram illustrates the framework's operational flow. The robot's observation consists of an RGB image, which is processed by the perception module to generate a segmented image using semantic segmentation. Based on the robot's traversability capabilities, navigable and non-navigable classes are defined. Utilizing this definition of navigability, a visual horizon is created to produce a navigability image. The robot then plans a visual path on this navigability image and employs visual servoing to navigate through the environment.}
    \label{fig:framework}
\end{figure*}

\subsection{Problem Formulation}
\label{sec:problem_formulation}
Consider a mobile robot operating in an \emph{unknown} environment $\mathbb{E} \subset \mathbb{R}^2$ where \emph{dynamic} obstacles are present. The robot's \emph{state} at time $t$ is defined by its pose $\mathbf{s}_t \triangleq [x_t, y_t, \theta_t]$, where $\mathbf{p}_{t}\triangleq[x_t, y_t]$ represents the position in the 2D plane (or locally approximated 2D surface), and $\theta_t \in [-\pi, \pi)$ denotes the heading angle.
The robot's \emph{observation} space consists of RGB images captured by an onboard camera, i.e., $\mathbf{I}_t \in \mathbb{R}^{H \times W \times 3}$, where $H$ and $W$ are the height and width of the image, respectively. 
The \emph{actions} or \emph{control commands} that the robot can execute $\mathbf{a}_t = [v_t, \omega_t]$ are defined by the linear velocity $v_t$ and angular velocity $\omega_t$.

Given a goal position $\mathbf{g} = [x_g, y_g] \in \mathbb{E}$ specified in the robot's initial coordinate frame, the objective is to move to the goal \emph{efficiently} while avoiding \emph{non-navigable areas}.
\begin{definition}[Non-Navigable Areas]
    Non-navigable areas $\mathbb{O}_{t}$ are defined as regions where the robot cannot traverse due to obstacles, safety constraints, physical limitations, or other user-defined criteria for selective navigation.
\end{definition}
The visual navigation problem can be formulated as finding a \emph{policy} $\pi$, mapping the visual observation to control commands:
\begin{equation}
    \mathbf{a}_t = \pi(\mathbf{I}_t).
\end{equation}

The robot's motion is subject to the following constraints:
\begin{align*}
    \mathbf{s}_{t+1} &= \mathtt{f}(\mathbf{s}_t, \mathbf{a}_t), && \text{(Motion Constraint)} \\
    \mathtt{d}(\mathbf{s}_t, \mathbb{O}_t) &\geq D_\text{safe}, && \text{(Safety Constraint)} \\
    |v_t| &\leq V_\text{max}, && \text{(Linear Velocity Bound)} \\
    |\omega_t| &\leq \Omega_\text{max}, && \text{(Angular Velocity Bound)} \\
    |v_t - v_{t-1}| &\leq \Delta V_\text{max}, && \text{(Linear Acceleration Bound)} \\
    |\omega_t - \omega_{t-1}| &\leq \Delta \Omega_\text{max}, && \text{(Angular Acceleration Bound)}
\end{align*}
where $\mathtt{f}(\cdot,\cdot)$ represents the unknown transition function related to the robot's dynamics, kinematics, and the environment, $\mathtt{d}(\mathbf{s}_t, \mathbb{O}_t)$ is the distance to the nearest non-navigable area, and other terms are self-explanatory.

The objective is to minimize the expected time to reach the goal while satisfying the above constraints:
\begin{equation}
    \min_{\pi} \mathtt{E}[T]\quad\text{ s.t. }\quad\norm{\mathbf{p}_{T} - \mathbf{g}}\leq\varepsilon,
\end{equation}
where $T$ is the time to reach the goal and $\epsilon$ is the acceptable goal region radius. When the robot's velocity is bounded, the time to reach the goal is a function of the path length and the robot's velocity profile; many existing works focus on optimizing the path length~\citep{anderson2018evaluation}. To simplify the notation, we will omit the time index $t$ in the following sections unless necessary.

%% file: 4_methodology.tex
\section{Methodology}\label{sec:methodology}

In this section, we present a comprehensive framework for mapless visual navigation that enables robots to navigate efficiently in unknown environments using only camera input. Our approach combines semantic segmentation, horizon-based planning, and visual servoing to achieve robust navigation performance without relying on prior maps or explicit 3D reconstruction.

\subsection{Framework Overview}

The proposed framework, POVNav, consists of two main components: a navigability image and a visual planner. The navigability image offers a simple and efficient representation of the intricate real-world environment. The visual planner addresses the competing navigation objectives by initially identifying a Pareto-optimal visual subgoal. The subsequent utilization of visual path planning facilitates the generation of navigation features that drives the robot through visual servoing, as depicted in ~\Cref{fig:framework}. We describe these critical modules in this section.

\subsection{Segmentation Algorithms}
To segment navigable regions in indoor environments, we adopt a self-developed surface normal-based segmentation method. For outdoor scenarios, we utilize the state-of-the-art semantic segmentation framework Mask2Former~\citep{cheng2022masked}. This section details the surface normal-based approach used for generating navigability images from depth data.

Let $\mathbf{I}_t^{\text{depth}} \in \mathbb{R}^{H \times W}$ denote the depth image at time $t$, where each pixel value $\mathbf{I}_t^{\text{depth}}(i,j)$ represents the depth measurement at image coordinates $(i,j)$. We assume that the depth image corresponds to a continuous 3D scene in the real world. To compute the surface normal at a given pixel, we locally approximate the surface as a triangle using three adjacent points: $\mathbf{I}_t^{\text{depth}}(i,j)$, $\mathbf{I}_t^{\text{depth}}(i-1,j)$, and $\mathbf{I}_t^{\text{depth}}(i,j-1)$. The corresponding 3D points are used to compute the surface normal via the cross product:
\begin{align}
    \overrightarrow{n}_{i,j} &= \left(\mathbf{I}_t^{\text{depth}}(i,j) - \mathbf{I}_t^{\text{depth}}(i-1,j)\right) \nonumber \\
    &\quad \times \left(\mathbf{I}_t^{\text{depth}}(i,j) - \mathbf{I}_t^{\text{depth}}(i,j-1)\right).
\end{align}

We then normalize this vector to obtain the unit surface normal:
\begin{equation}
    \widehat{n}_{i,j} = \frac{\overrightarrow{n}_{i,j}}{\| \overrightarrow{n}_{i,j} \|}.
\end{equation}

The set of all unit normals defines a surface normal image $\mathbf{I}_t^{\text{norm}} \in \mathbb{R}^{H \times W \times 3}$. To construct the binary navigability map $\mathbf{I}_t^{\text{nav}} \in \{0, 1\}^{H \times W}$, we use a geometric heuristic: a pixel is considered navigable if its normal vector points upward (i.e., aligned with the world $z$-axis):
\begin{equation}
    \mathbf{I}_t^{\text{nav}}(i,j) =
    \begin{cases}
        0, & \text{if } \widehat{n}_{i,j} \approx (0, 0, 1) \\
        1, & \text{otherwise}
    \end{cases}
\end{equation}

Here, $0$ denotes navigable (i.e., ground) and $1$ denotes non-navigable areas. In practice, a tolerance threshold is used to allow slight deviations from the exact vertical orientation due to surface irregularities and sensor noise.

Post-processing is applied to refine the segmentation: small navigable regions completely enclosed by non-navigable areas are reclassified as non-navigable, as they are unreachable. Additionally, any regions occluded behind non-navigable structures are also treated as non-navigable to ensure safe operation.
This method is lightweight, runs in real-time, and provides a robust geometric prior for ground segmentation in structured indoor environments.

Segmenting outdoor unstructured environments requires advanced methods. Transformer-based approaches have gained popularity, and we adopt Mask2Former~\citep{cheng2022masked} for this purpose due to its excellent performance. Additional technical details can be found in the original paper.

\subsection{Start and Goal in Image Space}
In traditional robot navigation, planning occurs in real-world coordinates using a predefined map. However, our problem is different: the robot only knows the relative position of its goal (with respect to its current position) and must navigate without a map, planning its path directly in the image space. A key challenge in image-space planning is determining the start and goal positions in the image.

\begin{figure}[t!]
    \begin{center}
        \includegraphics[width=\linewidth]{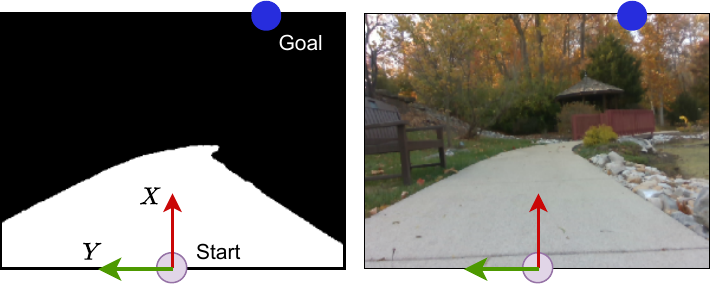}
    \end{center}
    \caption{The planning image and its corresponding input image are shown. The reference frame is fixed at the bottom center whose $x$-axis is shown by red arrow and $y$-axis is shown by green arrow.}\label{fig:planning_space}
\end{figure}

\begin{figure}[t!]
    \centering
    \subfigure[Forward Goal]{
        \includegraphics[width=3.9cm]{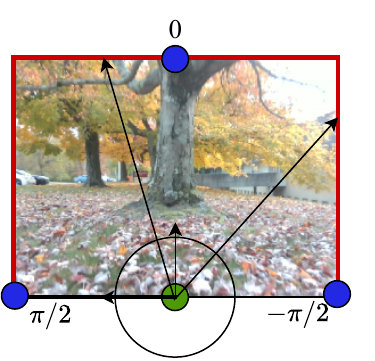}
    \label{fig:POG1}}
    \subfigure[Backward Goal]{
        \includegraphics[width=3.9cm]{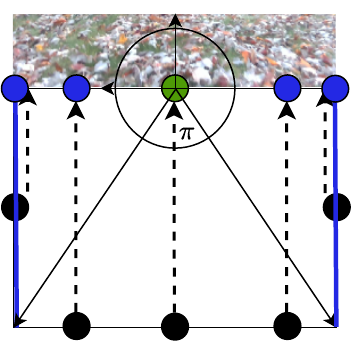}
    \label{fig:POG2}}
    \caption{Illustration of the goal angle mapping on the image border. (a) Visualizes the mapping of goals located in front of the robot, corresponding to angles $[-\pi/2, \pi/2]$, onto the left, top, and right borders of the image. Blue circles indicate sample POG points for the goal directions at $-\pi/2$, $0$, and $\pi/2$, with the green circle denoting the robot's current position. The two arrows illustrate example mappings for representative goal directions located to the left and right of the robot’s heading. (b) Illustrates how goals located behind the robot, with angles in $(\pi/2, \pi]$ and $(-\pi/2, -\pi)$, are projected onto an extended virtual border and then re-mapped to the bottom edge using a dashed projection. 
    Black circles represent virtual goal locations, while blue circles denote actual goal projections.}
    \label{fig:pogoalg_mapping}
\end{figure}

The start position is typically chosen as the robot's current position. However, in the image space, the robot's current position is not directly observable. Computing the robot's current pose in the image is challenging due to the lack of visual cues. To simplify this process, we assume the robot's current position is at the \emph{bottom-center pixel} of the first-person-view image, as shown in \Cref{fig:planning_space}. This simplification is reasonable for a robot with a fixed camera setup that satisfies \Cref{assumption:camera}.

When the goal lies within the camera's field of view, we can compute its corresponding pixel location by transforming the relative goal position to image space, using parameters obtained during camera calibration. However, when the goal is outside the field of view, we must identify a \emph{surrogate goal} in the image.
To do so, we transform the relative goal direction (with \Cref{assumption:relative}) in the real-world to the \emph{image border} as they are the farthest observable pixels in the image. Specifically, we treat the bottom-center pixel (i.e., the start) as the origin of the image space, with the $x$-axis pointing upwards and the $y$-axis pointing to the left. To map goal angles to border pixels, we connect each border pixel to the origin with a line segment and calculate its angle relative to the $x$-axis. Through this mapping, angles in the range $[-\nicefrac{\pi}{2}, 0)$, 0, and $(0, \nicefrac{\pi}{2}]$ correspond to pixels along the right, up, and left borders of the image as shown in Figure~\ref{fig:POG1}. Note that the goal directions are defined with respect to the image's vertical axis ($x$-axis as shown in Figure~\ref{fig:planning_space}).
Figure~\ref{fig:POG2} illustrates the mapping of the remaining goal angles. Since the angles in the range $(\nicefrac{\pi}{2}, -\nicefrac{\pi}{2}]$ must all be mapped to a single available border—the bottom border, we mirror the left, top, and right borders with respect to the bottom edge. Using this mirrored structure, we apply the same mapping strategy to determine the surrogate goals for each pixel along the extended (virtual) border. However, because these virtual borders do not physically exist in the image, the corresponding points are projected back onto the real bottom border, as shown by the dashed arrows in Figure~\ref{fig:POG2}.

Interestingly, all goal directions lying on the mirrored left and right virtual borders (highlighted by the blue lines in the Figure~\ref{fig:POG2}) are projected to a single pixel each: the bottom-left corner for the left virtual border and the bottom-right corner for the right virtual border. While this creates a temporary saturation of surrogate goals at these corners, it does not impact navigation. As the robot rotates to align with the goal direction, these angles gradually move into the visible region, updating their projections accordingly.
We refer to this border-based surrogate goal as the {\em Peripheral Optic Goal (POG)}. It is worth mentionting that, although this is a practical and effective mapping, alternative strategies could also be considered depending on the application context.

\subsection{Navigability Image}
After defining the start and goal in the image space, we construct a \emph{navigability image} that indicates the navigable and non-navigable pixels in the image, which can be used for image-space planning.

\begin{definition}[Navigability Image]
    A navigability image $\mathbf{I}_{\text{nav}} \in \left\{ 0, 1 \right\}^{W\times{H}}$ is a binary image that represents the navigable and non-navigable areas in the \emph{image space}.
\end{definition}

One way to generate the navigability image is to use an off-the-shelf segmentation algorithm to classify each pixel into semantic classes. Specifically, at each time step, the segmentation model can take the observed image $\mathbf{I} \in \mathbb{R}^{H \times W \times 3}$ as the input and return a segmented image $\mathbf{I}_\text{seg} \in \mathbb{R}^{H \times W \times C}$, where $C$ is the number of segmentic classes. Based on the robot's specifications (e.g., dimensions, tire traction, motor power), a \emph{navigability function} is defined to classify each semantic class as navigable (denoted by $0$) or non-navigable (denoted by $1$). This function can also be learned from data collected during robot's past operation. Examples of navigable classes include grass, trail, and asphalt, while non-navigable classes include sky, tree, and building. The navigability image $\mathbf{I}_{\text{nav}}$ is then obtained by applying the navigability function to  each pixel in the segmented image $\mathbf{I}_\text{seg}$.

A straightforward idea is to run some path planning algorithms directly on the navigability image. However, planning on the navigability image can be challenging and unnecessary. Segmentation results obtained from real-world images are noisy, which can lead to incorrect navigability classification, especially in the upper part of the image where things are far away and appear small. Some scattered navigable spots in the mostly non-navigable region can mislead optimization-based planners to generate suboptimal or even infeasible paths. On the other hand, non-navigable spots in the mostly navigable region increase the difficulty of sampling-based planners. Due to this reason, sometimes there is no feasible path that connects the start and goal pixels within the navigable region. On top of that, spending computational resources on planning a complex path in the far-away region is unnecessary since it does not significantly affect the robot's immediate navigation decisions but only provides a rough direction to the goal. 

\subsection{Visual Horizon}

\begin{figure}[t]
    \centering
    \includegraphics[width=0.9\linewidth]{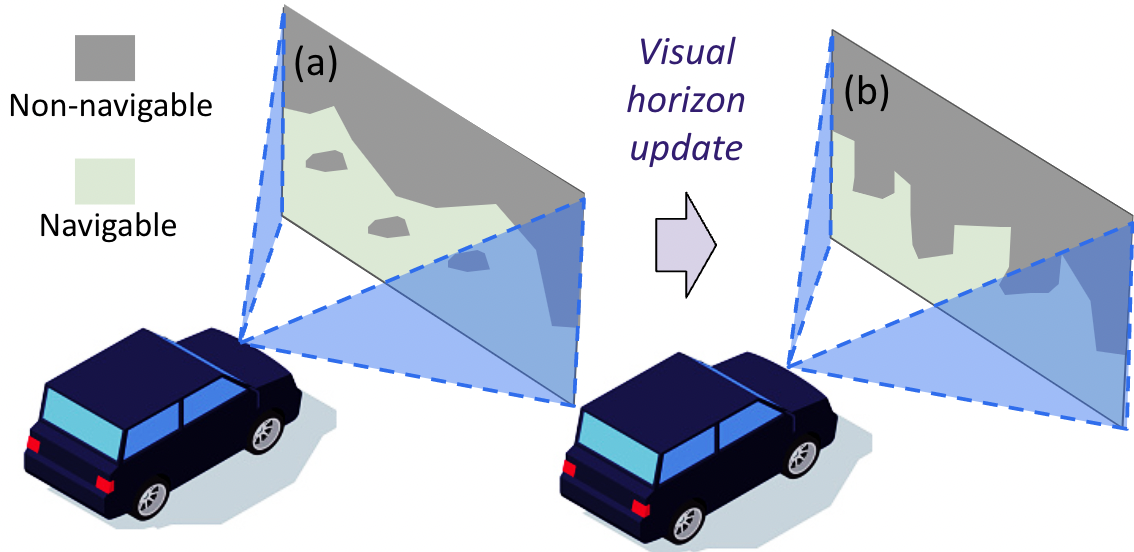}
    \caption{An illustration of the raw and processed navigability image.}\label{fig:horizon_update}
\end{figure}

To simplify the planning problem, we introduce the concept of \emph{visual horizon}~\Cref{fig:horizon_update}. Intuitively, the visual horizon delineates the maximum extent of the navigable region surrounding the robot. By focusing on the region within the horizon, we can simplify the planning problem and more confidently generate a path that is traversable by the robot.

\begin{definition}[Visual Horizon]
    The visual horizon $\mathbb{H}$ is a set of \emph{contiguous} pixels of the navigability image $\mathbf{I}_{\text{nav}}$. The upper boundary of the visual horizon starts from the left border and ends at the right border such that (a) any pixel below the horizon is navigable; (b) the pixels on and above the horizon are non-navigable; and (c) the number of navigable pixels below the horizon is maximized. In addition, the left and right image borders below the horizon are also considered part of the visual horizon.
\end{definition}

In practice, the visual horizon boundary can be computed by scanning the navigability image from left to right. The basic idea is to initialize the visual horizon to be the bottom row of the image and ``push'' each pixel on the horizon {upwards} until it reaches a non-navigable pixel. This process is repeated for each column of the image. 

The visual horizon provides a clear and contiguous boundary for the navigable region surrounding the robot. Now we can bypass the aforementioned challenges of planning on the navigability image by planning \emph{below} the visual horizon.
Noticing that the POG is typically outside the visual horizon, we need to find a \emph{subgoal} within the visual horizon for planning. Since any pixel below the horizon is navigable, the planning problem degenerates into a naive problem of drawing a straight line from the start to the POG. Any point where the line segment connecting the start to the POG intersects the horizon can serve as a subgoal. However, this straightforward approach can lead to suboptimal paths where the robot keeps moving towards obstacles.
To address this issue, we introduce a subgoal selection mechanism. 
\subsection{Pareto-Optimal Subgoal Selection}

Selecting a subgoal is essentially deciding the optimization criteria for the robot's immediate navigation decision which might involve multiple conflicting objectives.

\subsubsection{Objective Functions:}

The naive approach is to {minimize the distance to the goal}, yielding a shortest path. To avoid bumping into obstacles, it also needs to add another optimization criterion that {maximizes the clearance from obstacles}. However, these two objectives are typically conflicting: moving towards the goal often means maneuvering closer around obstacles when the goal is behind an obstacle and thus leads to a collision risk. In addition, tunning the weights of these two objectives is non-trivial and can be time-consuming. In many scenarios, the robot may also get stuck in local minima.

Our insight is that the robot does not need to move directly {towards} the goal. Instead, it only needs to move as far as possible from the current location while getting closer to the goal and avoiding non-navigable areas. This insight leads to the formulation of the subgoal selection problem as a {\em multi-objective} optimization problem.

We prioritize the safety of the robot over the other two objectives, so the traversability objective is not implemented as an optimization criterion. Instead, we implicitly optimize the traversability objective by constraining the subgoal to lie within the visual horizon.

\begin{definition}[Traversability Objective]
    The search space of the subgoal $\mathbb{G}$ is constrained to the pixels \emph{on} or \emph{below} the visual horizon such that there exists a path connecting the start and the subgoal that is entirely within the navigable region. Formally, for each pixel $(x, y) \in \mathbb{G}$,
    \begin{equation}
        \forall (x, y) \in \mathbb{G} \quad \exists (x, y') \in \mathbb{H} \quad \text{ s.t. } y' \geq y.
    \end{equation}
\end{definition}

\begin{definition}[Navigation Objective]
    Given the coordinates of the peripheral optic goal $(x_{g}, y_{g})$, the navigation objective is to find a subgoal $(x, y) \in \mathbb{G}$ that minimizes the deviation $\mathtt{c_\text{nav}}$ of the angle  pointing towards the subgoal $\theta_{s}\triangleq\mathtt{arctan}(\nicefrac{y}{x})$ from the goal angle $\theta_{g}\triangleq\mathtt{arctan}(\nicefrac{y_{g}}{x_{g}})$:
    \begin{equation}
        \argmin_{(x, y)}\mathtt{c_\text{nav}}(x, y) = \left| \theta_{s} - \theta_{g} \right|.
        \label{eq:nav_obj}
    \end{equation}
\end{definition}

Since larger distance between two pixels in the image space corresponds to larger distance of the two locations in the real world, we can simply use the Euclidean distance between the subgoal and the start pixel as the exploration objective formally defined as follows.

\begin{definition}[Exploration Objective]
    The exploration objective is to maximize the distance from the current location (i.e., the origin) to the subgoal $(x, y)$ or equivalently, to minimize the following cost function:
    \begin{equation}
        \argmin_{(x, y)}\mathtt{c_\text{exp}}(x, y) = -\norm{(x, y)}.
        \label{eq:exp_obj}
    \end{equation}
\end{definition}

\subsubsection{Optimality and Search Space:}

These two objectives are often conflicting in nature.
Each candidate subgoal $(x, y)$ is associated with a vector of objective values $\mathbf{c}$. For example, here we have defined two objectives $\mathbf{c}\triangleq[\mathtt{c_\text{nav}}(x, y), \mathtt{c_\text{exp}}(x, y)]$, which can be further extended to a multi-objective optimization problem with more objectives.
Let $c_d$ be the $d$-th element of $\mathbf{c}$.
\begin{definition}[Pareto Optimal Solutions]
    A subgoal $(x, y)$ is better than, or \emph{dominating}, another subgoal $(x', y')$ if and only if the following conditions are satisfied:
    \begin{itemize}
        \item Any element of $\mathbf{c}$ is not larger than the corresponding element in $\mathbf{c}'$: $\forall d, c_d \leq c'_d$;
        \item At least one element of $\mathbf{c}$ is smaller than the corresponding element in $\mathbf{c}'$: $\exists d \text{ s.t. } c_d < c'_d$.
    \end{itemize}
    The set of solutions that are not dominated by any other solutions is called the \emph{Pareto-optimal set}.
\end{definition}

All subgoals in the Pareto-optimal set are \emph{equally good} in terms of the objectives and the subgoal should be selected \emph{only} from this set.

\begin{lemma}[Optimal Subgoal on Visual Horizon]\label{lemma:optimal_subgoal}
    The subgoal $(x^\star, y^\star)$ that is Pareto-optimal with respect to the navigation and exploration objectives must lie on the visual horizon $\mathbb{H}$.
\end{lemma}
\begin{proof}
    This can be proved by contradiction. Assume that there is a Pareto-optimal subgoal $(x^\star, y^\star)$ that does not lie on the visual horizon. Draw a line starting at the robot's current position $(0, 0)$ and passing through $(x^\star, y^\star)$. Since $(0, 0)$ and $(x^\star, y^\star)$ are both \emph{within} the visual horizon, this line must intersect the visual horizon at some point $(x', y')$. This point has the same navigation objective value as $(x^\star, y^\star)$ but a better exploration objective value, dominating $(x^\star, y^\star)$, hence contradicting the assumption that $(x^\star, y^\star)$ is Pareto-optimal.
\end{proof}

\Cref{lemma:optimal_subgoal} greatly simplifies subgoal selection by restricting the search space to the visual horizon $\mathbb{H}$. Since $\mathbb{H}$ can be \emph{enumerated} efficiently, this bypasses the need for complex optimization algorithms. We introduce the term Horizon Optic Goal (HOG) to denote the subgoal selected from the Pareto-optimal set on the visual horizon, distinguishing it from the POG (the goal projected to the image border).

\subsubsection{Horizon Optic Goal Selection:}

The remaining challenge is to pick a single HOG from the horizon. We choose a \emph{scalarization} approach to convert the multi-objective optimization problem into a single-objective optimization problem by weighting the objectives. The scalarization function is defined as:
\begin{equation}
    \mathtt{c}(x, y) = w_1 \mathtt{c_\text{nav}}(x, y) + w_2 \mathtt{c_\text{exp}}(x, y),
    \label{eq:scalarization}
\end{equation}
where $w_1$ and $w_2$ are the weights for the navigation and exploration objectives, respectively. The weights can be set based on the complexity of the environment and/or the user preferences. For example, if the robot is in a cluttered environment, the weight for the exploration objective should be higher to avoid obstacles.

We adopt the scalarization approach because it is simple, computationally efficient, and provides a clear trade-off between the two objectives. However, there are two caveats to be aware of. First, scalariation cannot find \emph{all} Pareto-optimal solutions if the Pareto front is non-convex. 
In our case, the HOG only needs to be \emph{good enough} for the robot to make the current navigation decision so finding a single compromised solution efficiently is preferable to exploring the full trade-off space. Second, if the objectives are not naturally comparable in scale, the weights $w_1$ and $w_2$ need to be carefully tuned or the objectives need to be normalized.

\subsection{Efficient and Effective Servoing Features}

\begin{figure}[tbp]
    \centering
    \includegraphics[width=.6\linewidth]{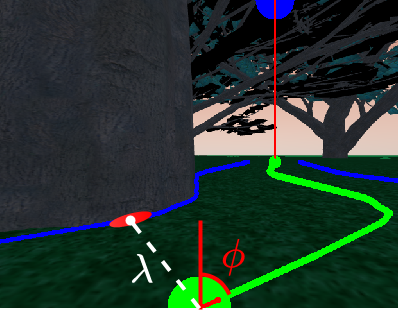}
    \caption{An illustration of a visual path (green line) planned from the start position (green disc) to the Horizon Optic Goal (blue disc) with a safe boundary (blue lines) that account for the size of the robot and a safety margin.}\label{fig:feature_gen}
\end{figure}

After selecting the HOG, the robot needs to take actions to move towards this visual subgoal -- a process known as {visual servoing}~\citep{chaumette2006visual}. The robot's motion is controlled based on visual features extracted from the image. The visual features should be informative enough to guide the robot towards the HOG while driving defensively to avoid non-navigable areas. Although the HOG is on the visual horizon of the current image and all pixels below the horizon are navigable, the robot should not rush directly towards the HOG because the obstacles can be dynamic and the segmentation boundaries are noisy. To address this issue, we propose to generate two features: the \emph{proximity} feature and the \emph{alignment} feature.

The proximity feature measures the distance from the robot to the closest pixel on the visual horizon.
\begin{definition}[Proximity Feature]
    The proximity feature $\lambda$ is defined as the minimum Euclidean distance from the robot's current position to the visual horizon $\mathbb{H}$:
    \begin{equation}
        \lambda = \min_{(x, y) \in \mathbb{H}} \norm{(x, y)}.
    \end{equation}
\end{definition}
The proximity feature is used to control the robot's linear velocity. Intuitively, the robot should move faster when the proximity is large and slow down when the proximity is small.

The alignment feature measures the deviation of the robot's heading direction from the direction pointing towards a safe path connecting the robot to the HOG. In \Cref{fig:feature_gen}, the heading direction is represented by the red thick line and the planned path is represented in the green line.

\begin{definition}[Alignment Feature]
    Let $\boldsymbol{\xi}$ be a sequence of pixels representing a safe path connecting the robot to the HOG.
    Draw a circle centered at the robot's current position $(0, 0)$ with radius $R$ such that the circle intersects the safe path $\boldsymbol{\xi}$ at a single pixel $(x_*, y_*)$. The alignment feature $\phi$ is defined as the angle between the robot's heading direction and the direction pointing towards $(x_*, y_*)$:
    \begin{equation}
        \phi = \mathtt{arctan}\left(\frac{y_*}{x_*}\right).
    \end{equation}
\end{definition}
The alignment feature is used to control the robot's angular velocity. The robot adjust its heading direction to align with the safe path.

Consider the robot's state defined by the proximity feature $\lambda$ and alignment feature $\phi$. Let $\lambda^*$ represent the desired safe distance from obstacles and $\phi^* = 0$ represent perfect alignment with the safe path. We propose the following control law: 
\begin{equation}
\begin{aligned}
v &= k_v(\lambda - \lambda^*), \\
\omega &= -k_\omega\phi,
\end{aligned}
\end{equation}
where $v$ is the linear velocity, $\omega$ is the angular velocity, and $k_v, k_\omega > 0$ are control gains.

To ensure that the control commands respect the physical and operational constraints defined in the problem formulation (Section~\ref{sec:problem_formulation}), we explicitly bound the control inputs as:
\[
v \in [-V_{\text{max}}, V_{\text{max}}], \quad \omega \in [-\Omega_{\text{max}}, \Omega_{\text{max}}],
\]
where $V_{\text{max}}$ and $\Omega_{\text{max}}$ denote the platform-specific limits on linear and angular velocities. These bounds reflect actuation capabilities, mechanical constraints, and safety requirements of the robot.

Accordingly, we revise the control law to enforce these limits using nested $\min$ and $\max$ operations:
\begin{equation}
\begin{aligned}
v &= \max\left(-V_{\text{max}}, \min\left(k_v(\lambda - \lambda^*), V_{\text{max}}\right)\right), \\
\omega &= \max\left(-\Omega_{\text{max}}, \min\left(-k_\omega \phi, \Omega_{\text{max}}\right)\right).
\end{aligned}
\end{equation}
This formulation ensures that the control commands $\mathbf{a} = [v, \omega]$ always lie within the feasible set specified in the problem constraints.

In high-speed or traction-limited scenarios, coupling between translational and rotational motions may induce excessive lateral forces. Specifically, the lateral acceleration incurred during turning is $a_{\text{lat}} = v \omega$, which—combined with longitudinal acceleration $a_{\text{long}}$, must satisfy the traction constraint defined by the friction circle 
$\sqrt{a_{\text{lat}}^2 + a_{\text{long}}^2} \leq \mu g$,
where $\mu$ is the surface friction coefficient and $g$ is gravitational acceleration. To remain within this constraint, the translational velocity $v$ can be adaptively modulated based on the angular velocity $\omega$ and estimated surface traction. While this coupling is not critical in our current experiments which operate at moderate speeds, the proposed formulation can be extended to support traction-aware control for high-speed navigation or uneven terrain where dynamic feasibility becomes a dominant concern.

\begin{figure*}[t!]
    \centering
    \subfigure[Normal Situation]{
        \includegraphics[width=3.9cm]{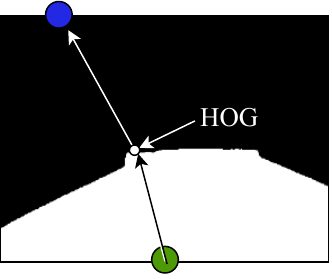}
    \label{fig:vh1}}
    \subfigure[Downhill]{
        \includegraphics[width=3.9cm]{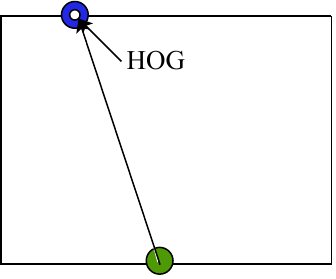}
    \label{fig:vh2}}
    \subfigure[Uphill]{
        \includegraphics[width=3.9cm]{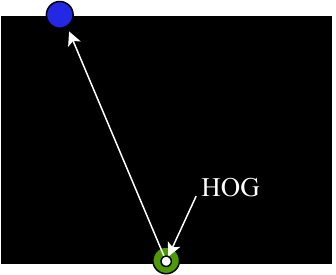}
    \label{fig:vh3}}
    \subfigure[Narrow Passage]{
        \includegraphics[width=3.9cm]{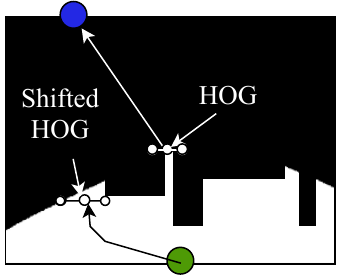}
    \label{fig:vh4}}
    \caption{Illustration of various scenarios where the visual horizon influences path planning and visual servoing (blue circle shows the POG and green circle shows the start position in all the subfigures). (a) A typical flat-ground navigation case, where the HOG lies on the visual horizon, which cleanly separates navigable and non-navigable regions. (b) A downhill navigation scenario, where the robot sees only navigable terrain, shifting the visual horizon to the image borders and moving the HOG to the POG. (c) An uphill traversal, where the robot encounters only non-navigable regions, pushing the visual horizon to the bottom border of the image and relocating the HOG to the start position. (d) A constrained passage case, where a safe path to the HOG cannot be generated due to a narrow gap. The HOG is shifted to a nearby safe region, resulting in an incomplete path. The horizontal line segments shown here at the HOG indicate the projected robot footprint in image space, which increases in size as the robot approaches the start position. }
    \label{fig:visual_horizon_special_cases}
\end{figure*}

\begin{algorithm}[t!]
    \setstretch{1.1}\small
    \caption{\textbf{Visual Path Generation}}
    \label{alg:safe_path}
    \begin{algorithmic}[1] 
        \Require $\mathbf{I}_{\text{nav}}$, $(x^\star, y^\star)$
        \Ensure Path $\xi$ and $\texttt{mode}$
        \Comment{$\texttt{mode} \in \{\texttt{safe}, \texttt{fallback}\}$}
        \State Initialize $(x, y) \gets (x^\star, y^\star)$ \Comment{Start from HOG}
        \State Set $\texttt{shift\_direction} \gets \text{NULL}$, \texttt{safe\_margin} $\gets$ constant
        \State Initialize empty path $\xi \gets [\ ]$ and $\texttt{mode} \gets \texttt{safe}$
        \While{$y \geq 0$}
            \State $\Delta x \gets \texttt{find\_safe\_distance\_in\_image\_space}(y)$
            \State $x_{\text{left}} \gets x + \Delta x$
            \State $x_{\text{right}} \gets x - \Delta x$
            \If{any pixel on $[x_{\text{left}}, x_{\text{right}}]$ is non-navigable}
                \If{$\texttt{shift\_direction} = \text{NULL}$}
                    \State $\texttt{shift\_direction} \gets \texttt{find\_shift\_direction()}$
                    \If{$\texttt{shift\_direction} = \text{NULL}$}
                        \State $\texttt{mode} \gets \texttt{fallback}$
                        \State $y \gets y - 1$
                        \State Shift $x$ towards the start point direction
                        \State \textbf{continue} \Comment{Skip adding this point and move to next row}
                    \EndIf
                \EndIf
                \State Shift $x$ in the $\texttt{shift\_direction}$ until all points on $[x_{\text{left}}, x_{\text{right}}]$ are navigable
            \EndIf
            \State Append $(x, y)$ to $\xi$
            \State $y \gets y - 1$
            \State Shift $x$ towards the start point direction
        \EndWhile
        \State Append $(0, 0)$ to $\xi$
        \State \Call{Reverse}{$\xi$}
        \State \Return $\xi$, $\texttt{mode}$
    \end{algorithmic}    
\end{algorithm}

To generate the visual path $\boldsymbol{\xi}$, any efficient path planning algorithm can be used. We propose a simple algorithm that generates a path connecting the robot to the HOG while considering the robot's size and a safety margin. The algorithm in \Cref{alg:safe_path} outlines the detailed process. Calculating the path in this order ensures feasibility within a linear time constraint. At each step, a safe distance is maintained on both sides of the path in image space. This safe distance is computed by adding a safety margin to half the robot's width and converting this value to image coordinates using the camera's intrinsic and extrinsic parameters, which include the current pixel position $y$, focal length $f$, real-world safety margin $R_{\text{safety\_margin}}$, and camera height from the ground $h$. 
A shift direction is then determined to move the path away from obstacles, ensuring a safe $x$-distance on either side of each point along the $y$-axis. If no shift direction is found, the algorithm marks that segment as unsafe and resumes checking lower rows. This results in either a fully safe path or a partial fallback trajectory. Note that the shift direction only needs to be identified once, as moving from the visual horizon or any navigable point to the starting point guarantees an obstacle-free path in this direction, due to the horizon's inherent structure.  

Figure~\ref{fig:visual_horizon_special_cases} illustrates several important edge cases in how the visual horizon affects planning and control.

In standard navigation scenarios (Figure~\ref{fig:vh1}), the visual horizon effectively separates navigable and non-navigable regions in the image. HOG lies on the visual horizon, and a safe, obstacle-free path from the HOG to the robot’s current position can be generated. In such cases, Algorithm~\ref{alg:safe_path} successfully returns a safe path.

However, real-world especially off-road environments often introduce more complex scenarios such as steep terrain or perception noise. For instance, when the robot is descending a slope (Figure~\ref{fig:vh2}), the camera may observe only navigable terrain. A similar situation may arise due to misclassification by the segmentation model, which could label all regions as navigable. In such cases, the HOG is shifted to coincide with the POG, resembling free-space navigation. A valid path can still be computed in this case.

Conversely, during uphill traversal (Figure~\ref{fig:vh3}), the camera may observe only non-navigable regions, or the segmentation model may erroneously classify the entire image as non-navigable, even when navigating a flat surface. In such cases, the HOG collapses to the start point, resulting in a path of zero length. This presents a unique challenge: the proximity feature drops to zero (triggering a backward motion), while the alignment feature becomes undefined due to the lack of a visible path. As a fallback, we align the robot to the POG to maintain a consistent heading.
However, the robot still perceives itself as being dangerously close to an obstacle and continues to move backward, which prevents it from climbing steep slopes. Alternative strategies such as retaining the last known safe proximity value or using a fixed fallback proximity could be used, but they risk causing collisions in ambiguous situations, such as when the robot is genuinely near an obstacle and cannot observe any navigable area.
This is a key limitation of the current method (and existing solutions). Addressing it would require incorporating additional information, such as camera orientation or fallback policies conditioned on pitch, potentially through the use of a gimbaled camera system or dynamic viewpoint adjustment. We plan to investigate these enhancements in future work to further improve system robustness. 

Another special case arises when the goal lies directly behind the robot ($\theta = \pi$). In this case, the start point, HOG, and POG may coincide, causing the alignment feature to become ill-defined. To prevent this, we introduce a small angular bias in the goal projection, ensuring numerical separation and allowing feature computation.

Finally, Figure~\ref{fig:vh4} shows a common off-road scenario in which thin tree branches or narrow passages obscure parts of the horizon, making it impossible to generate a fully safe path. In such cases, the proposed algorithm handles the failure by producing a fallback path that is partially safe, guiding the robot through available gaps.

\subsection{Theoretical Analysis}

We present theoretical analysis of the POVNav framework focusing on three key aspects: (1) optimality of the sub-goal selection, (2) control stability analysis, and (3) computational complexity analysis.

\subsubsection{Optimality Analysis:}

First, we establish that our scalarization-based HOG selection approach yields weakly Pareto-optimal solutions regardless of the convexity of the objective functions.

\begin{definition}[Weak Pareto Optimality]
A solution $(x^*, y^*)$ is weakly Pareto-optimal if there does not exist another feasible solution $(x, y)$ such that:
\[
c_i(x, y) < c_i(x^*, y^*) \quad \forall i \in \{1,\ldots,M\},
\]
where $c_i$ are the objective functions and $M$ is the number of objectives.
\end{definition}

\begin{theorem}[Weak Pareto Optimality of HOG]
Let $\mathbb{H}$ be the visual horizon and $(x^*, y^*)$ be the selected HOG that minimizes the scalarized objective:
\[
c(x,y) = w_1c_\text{nav}(x,y) + w_2c_\text{exp}(x,y),
\]
where $w_1, w_2 > 0$. Then $(x^*, y^*)$ is weakly Pareto-optimal.
\end{theorem}

\begin{proof}
We proceed by contradiction. Assume $(x^*, y^*)$ is not weakly Pareto-optimal. Then $\exists (x', y') \in \mathbb{H}$ such that:
\[
c_\text{nav}' < c_\text{nav} ^*\quad \text{and} \quad
c_\text{exp}' < c_\text{exp}^*.
\]
Here, we use the shorthand $c_\text{nav}' \triangleq c_\text{nav}(x', y'), c_\text{nav}^* \triangleq c_\text{nav}(x^*, y^*), c_\text{exp}^* = c_\text{exp}(x^*, y^*), \text{ and  } c_\text{exp}' = c_\text{exp}(x', y')$. Since $w_1, w_2 > 0$, we have:

\[
    w_1c_\text{nav}' + w_2c_\text{exp}' < w_1c_\text{nav}^* + w_2c_\text{exp}^*.
\]

This contradicts the assumption that $(x^*, y^*)$ minimizes $c(x,y)$. Therefore, $(x^*, y^*)$ must be weakly Pareto-optimal.
\end{proof}

\subsubsection{Control Stability Analysis:}

We now analyze the stability properties of the proposed visual servoing control law and establish conditions under which the robot will converge to its desired state.

\begin{definition}[Error State]
Let $\mathbf{e} = [\lambda - \lambda^*, \phi]^\top$ be the error state vector, where $\lambda - \lambda^*$ represents the proximity error and $\phi$ represents the alignment error.
\end{definition}

\begin{theorem}[Local Asymptotic Stability]\label{theorem:stability}
For sufficiently small gains $k_v, k_\omega > 0$, the control law
\begin{equation}
\begin{aligned}
v &= k_v(\lambda - \lambda^*) \\
\omega &= -k_\omega\phi
\end{aligned}
\end{equation}
ensures local asymptotic stability of the equilibrium point $\mathbf{e} = \mathbf{0}$.
\end{theorem}

\begin{proof}
Consider the Lyapunov function candidate:
\begin{equation}
V(\mathbf{e}) = \frac{1}{2}(\lambda - \lambda^*)^2 + \frac{1}{2}\phi^2.
\end{equation}

Taking the time derivative:
\begin{equation}
\dot{V}(\mathbf{e}) = (\lambda - \lambda^*)\dot{\lambda} + \phi\dot{\phi}.
\end{equation}

The feature dynamics can be expressed using the interaction matrix $\mathbf{L}$:
\begin{equation}
\begin{bmatrix} \dot{\lambda} \\ \dot{\phi} \end{bmatrix} = 
\begin{bmatrix} 
L_{11} & L_{12} \\
L_{21} & L_{22}
\end{bmatrix}
\begin{bmatrix} v \\ \omega \end{bmatrix}.
\end{equation}

Substituting the control law:
\begin{equation}
\dot{V}(\mathbf{e}) = -k_v(\lambda - \lambda^*)^2L_{11} - k_\omega\phi^2L_{22} + \text{cross terms}.
\end{equation}

For sufficiently small gains, the cross terms can be dominated by the diagonal terms. Since linear and angular velocities have positive correlation with the feature dynamics $L_{11}, L_{22} > 0$, we have $\dot{V}(\mathbf{e}) < 0$ for $\mathbf{e} \neq \mathbf{0}$. Therefore, by Lyapunov's direct method, the equilibrium point is locally asymptotically stable.
\end{proof}

\subsubsection{Complexity Analysis}

\begin{remark}[Time Complexity]
Given an image of size $W \times H$, the computational complexity of POVNav for one navigation step, excluding the segmentation step, is $O(WH)$.
\end{remark}
Let us analyze each major component:
\begin{itemize}
    \item Segmentation (External Module): A semantic segmentation model (e.g., ~\citep{chen2022cali, cheng2022masked} or lightweight rule-based methods) is used to assign class labels to each pixel. The complexity and runtime of this module depend on the specific model used and are considered external to the core POVNav system.
    
    \item Navigability Image Generation: Once semantic segmentation is available, each pixel is classified as navigable or non-navigable using a \emph{navigability function} that maps semantic classes to binary labels. This function may rely on a static lookup table defined by human input or a dynamic table generated using an LLM. Importantly, updates to this lookup table can be performed asynchronously and do not affect the real-time decision loop of POVNav. The binary segmentation lookup and visual horizon extraction together require $O(WH)$.
    
    \item Pareto-Optimal Subgoal Selection: This step involves scanning across the visual horizon to select a subgoal using scalarized multi-objective optimization, which takes $O(W)$.
    
    \item Feature Generation: The proximity feature is computed over horizon width ($O(W)$), and the alignment feature requires selecting a point from the path ($O(H)$).
    
    \item Control Command Generation: A bounded control law maps features to velocity commands in constant time, i.e., $O(1)$.
\end{itemize}
Since the segmentation method is modular and can be selected based on hardware constraints, its complexity is not included in the core complexity of POVNav. 

The total complexity is dominated by the navigability image generation and is $O(WH)$.

\begin{remark}[Space Complexity]
The space complexity of POVNav is $O(WH)$.
\end{remark}
The space requirements are:
\begin{itemize}
    \item Navigability image storage: $O(WH)$
    \item Visual horizon storage: $O(W)$
    \item Path and feature storage: $O(W)$
\end{itemize}
Thus the total space complexity is dominated by the image storage: $O(WH)$.

\subsection{Implementation Details}
Implementation of our framework requires careful consideration of several practical aspects to ensure robust performance across different environments and scenarios. First, when projecting the goal to POG, we employ a smooth mapping function that prevents discontinuities at the image boundaries. This helps maintain stable robot behavior when the goal transitions between different regions of the image border.

For navigability image generation, we utilized various methods: learning-based algorithms, depth-based surface normal segmentation, and color-based detection. Of these, color-based detection is lightweight enough to be implemented on a microcontroller, while depth-based surface normal segmentation can run on low-end devices like a Raspberry Pi without GPU support. The learning-based methods, which require GPU acceleration, are executed on a portable GPU machine compatible with small robots like the Jackal and Unitree Go1, as shown in \Cref{fig:exp_setup_robots}.
To manage noisy segmentation, morphological operations are applied to remove isolated pixels and smooth region boundaries. We also limit the pixel adjustments to the average of recent movements, ensuring stability in segmentation outputs.

The weights in the scalarization function are normalized based on the characteristic scale of each objective. Specifically, we normalize the navigation objective by $\pi$ radians (maximum possible angle difference) and the exploration objective by the image diagonal length. This ensures that both objectives contribute meaningfully to the final decision regardless of image resolution or goal location.

The control gains for linear and angular velocities are adapted based on the robot's current state and environment characteristics. When the proximity feature indicates nearby obstacles, we reduce the linear velocity gain to promote more cautious movement. Similarly, the angular velocity gain is increased when sharp turns are required to follow the planned path.

The radius $R$ for computing the alignment feature is set dynamically based on the current proximity feature. We typically use $R = \min(\nicefrac{\lambda}{2}, d_{\text{max}})$, where $d_{\text{max}}$ is a maximum look-ahead distance. This ensures that the robot maintains a reasonable planning horizon while avoiding reactionary behavior in cluttered environments.

To achieve efficient implementation, we use integral image techniques for computing the proximity feature and maintain a spatial index of the visual horizon points. 
All critical computations are vectorized and optimized for cache efficiency.

%% file: 5_experiments.tex
\section{Experiments}
In this section, we evaluate the performance of the proposed approach across a variety of scenarios. These include both simulated and real-world environments, covering indoor and outdoor settings, structured and unstructured terrains, and different seasons such as winter and fall. The method is also tested with various segmentation techniques, under extreme conditions like after heavy snowfall at -7°C, forest trails, and long-range navigation tasks (on the kilometer scale) with different definitions of traversability. We also evaluate performance at various times of the day—morning, afternoon, and evening—to test the robustness of existing semantic segmentation methods under different lighting conditions and assess POVNav's adaptability to segmentation inconsistencies. Details of these experiments are discussed in the following subsections.

\subsection{Experimental Setup}
\subsubsection{Simulation:}
To evaluate the proposed method, we used the Clearpath Jackal robot equipped with an Intel Realsense camera in the ROS Gazebo simulation environment. The simulation was run on an Intel Alienware 17 laptop, configured with 32 GB of RAM and an 8 GB GPU, providing the necessary computational power for real-time processing. We implemented surface normal-based segmentation\footnote{\url{https://github.com/Dpushp/depth2surface_normals_seg}} and used it for all simulation experiments. 
The goal direction is determined by using the ground-truth robot pose and the goal pose in the world frame. This mimics the scenarios of perfect segmentation or reliable goal pose estimation. 

\begin{figure}[tbp]
    \centering
    \subfigure[Jackal]{
        \includegraphics[height=2.9cm, width=2.6cm]{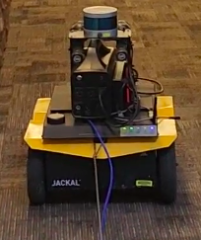}
    \label{fig:R1}}
    \subfigure[Husky]{
        \includegraphics[height=2.9cm, width=2.6cm]{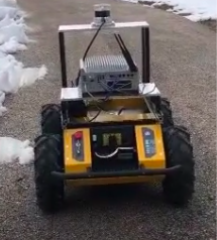}
    \label{fig:R2}}
    \subfigure[Unitree Go1]{
        \includegraphics[height=2.9cm, width=2.6cm]{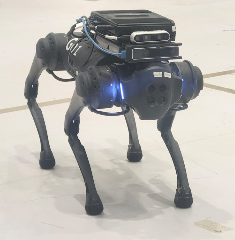}
    \label{fig:R3}}
    \caption{Robots used for real-world experiments. The Jackal robot was used for indoor navigation tasks and culvert inspection task, the Husky robot for outdoor navigation tasks, and the Unitree Go1 for both indoor and outdoor tasks.}\label{fig:exp_setup_robots}
\end{figure}
\subsubsection{Real-World:}
For real-world experiments, we utilized three different robotic platforms: Clearpath's Jackal, Husky, and the Unitree Go1 quadruped as shown in \Cref{fig:exp_setup_robots}. All robots were equipped with an Intel Realsense D435i camera for RGB-D sensing and an Intel Realsense T265 for localization (used to estimate the relative goal direction; this is not the only possible solution, as various other proprioceptive and exteroceptive sensors such as IMU, encoder, LiDAR etc can achieve the same objective). An Intel NUC with 32 GB of RAM and a 6 GB GPU was used as the processing unit across all platforms. This consistent hardware configuration ensured uniformity in performance evaluation across different platforms, allowing us to isolate the impact of environmental and task variations on the overall system performance. We use surface normal-based segmentation, an unsupervised learning-based method~\citep{chen2022cali} and Masked-attention Mask Transformer (Mask2Former)~\citep{cheng2022masked} for real-world experiments.

\subsection{Comparison with Baseline Methods}
We evaluate the performance of POVNav against two baseline methods across diverse simulated environments and tasks. The evaluation is conducted using key metrics: success rate, path length, and computation time per action. These metrics provide a comprehensive assessment of each method's efficiency and effectiveness, enabling a detailed comparison of POVNav's advantages and limitations relative to the baselines. The experiments are designed to rigorously test navigation robustness in both static and dynamic environments under varying conditions and task complexities.

\subsubsection{Baselines:} 
We compare the proposed POVNav method with a modified version of the Dynamic Window Approach (DWA)~\citep{fox1997dynamic}, which has been recently adapted to demonstrate the effectiveness of image segmentation for robotic navigation tasks~\citep{chen2022cali, chen2023polyline}. DWA is a widely adopted reactive planning method that shares navigation behavior similar to our framework. It restricts the search space to circular trajectories that are collision-free and reachable within a short time horizon.
The original DWA evaluates multiple trajectories based on predefined objective functions. In the approach proposed by \cite{chen2022cali, chen2023polyline}, motion primitives from DWA are projected onto the image, using a signed distance field for evaluation. Depth information is employed to project the candidate trajectories onto the images, allowing DWA to assess the proximity of obstacles. For collision checking, we assume that $\mathbb{H}$ represents an obstruction in the environment. We refer to this adapted method as the Visual Motion Primitive Planner (VMPP). For a fair comparison, we use our proposed {\em Navigability Image} as the local representation for this method. 

We also compare our method with a sparse optic flow-based reactive planner that utilizes time-to-transit ($\tau$) information~\citep{boretti2022ttt}. Inspired by biological systems, this approach uses monocular camera input to compute $\tau$ that is a visual cue derived from time-to-contact and applies it to steering control. The method employs robust vision-based navigation by leveraging $\tau$ values aggregated from features in the field of view. We refer to this method as SOFTNav (Sparse Optical Flow with Time-to-Transit Navigation) for showing the comparison in our evaluation.

To address concerns regarding robustness in dynamic environments, we additionally include two baseline methods from the modular and learning-based navigation paradigms.
As a representative modular approach, we evaluate our method against Model Predictive Contouring Control (MPCC)~\citep{mpcc}, which optimizes a sequence of control inputs to follow a reference path while accounting for dynamic constraints and obstacle avoidance. MPCC has demonstrated reliable performance in structured navigation tasks and allows us to assess the reactive capabilities of our method in the presence of dynamic pedestrians.
To show the advantage of proposed method over an end-to-end learning-based baseline, we compared our method with iPlanner~\citep{Yang-RSS-23}, a vision-based goal-conditioned local planner trained to produce steering and velocity commands directly from depth images and goal information. iPlanner has shown strong generalization in complex indoor and outdoor environments and serves as a strong baseline for comparing the robustness of our method to changes in appearance and scene layout.

\begin{figure*}[tbp]
    \centering
    \subfigure[Static Environment (Env1 to Env5)]{
        \label{fig:static_env}
    \includegraphics[width=2.2in]{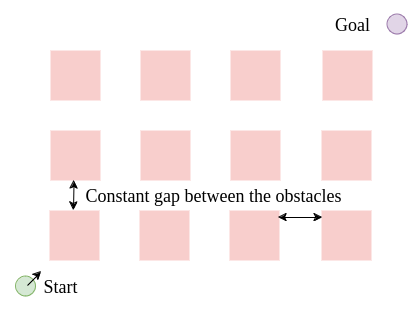}} 
    \subfigure[Dynamic Environment (Env6)]{
        \label{fig:dynamic_env}
    \includegraphics[width=4.35in]{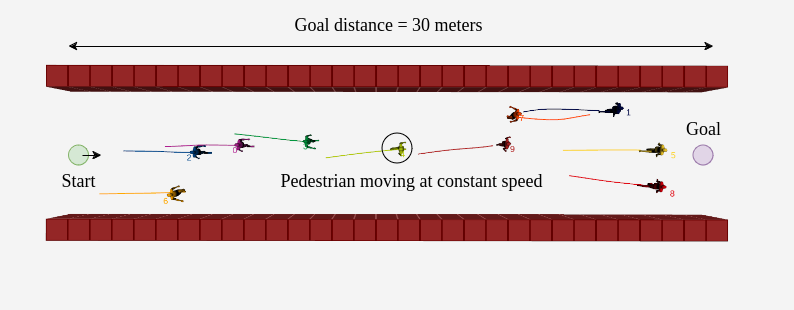}}
    \caption{(a) Illustrates the general layout of the simulated environment with static obstacles used for evaluating the static obstacle avoidance task. (b) Layout of the corridor environment with 10 pedestrians for testing dynamic obstacle avoidance. Pedestrians in the corridor move at different constant velocities, navigating based on the social force model. 
    }\label{fig:three graphs}  
\end{figure*}

\begin{figure}[tbp]
    \centering
    \includegraphics[height=3.0in]{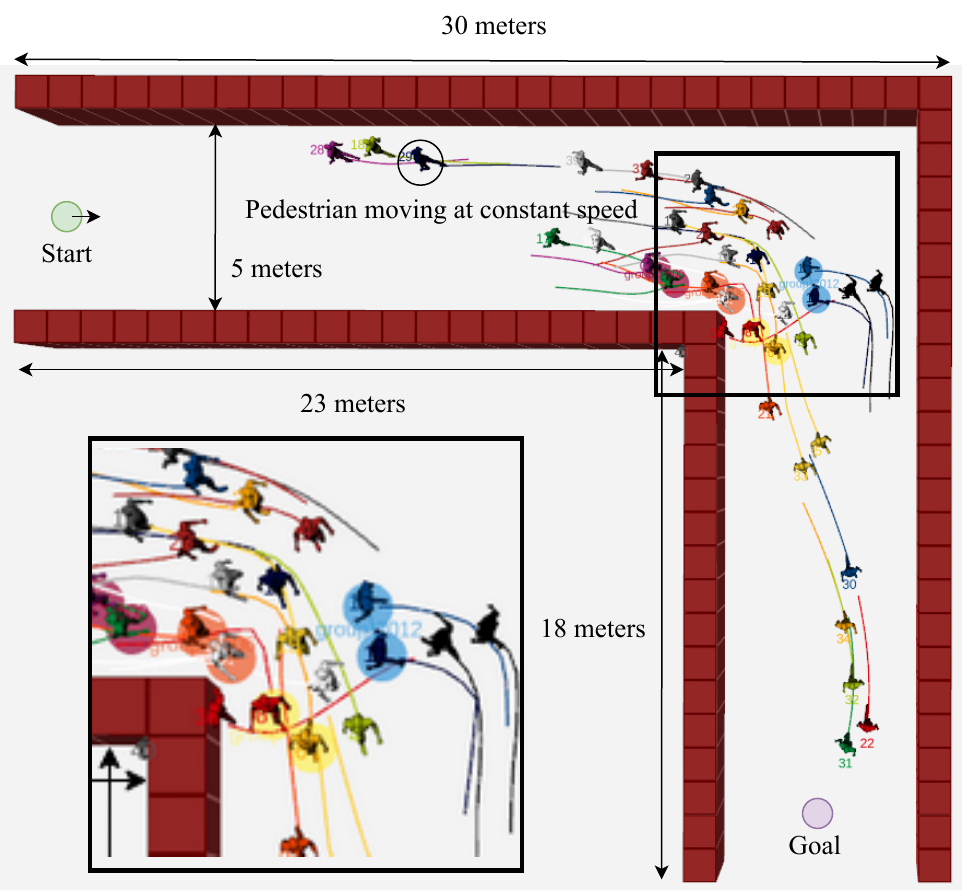}
    \caption{Layout of a more challenging environment: an L-shaped corridor populated with 40 pedestrians moving bidirectionally to simulate real-world dynamics. The zoomed-in inset at the lower left highlights diverse pedestrian behaviors and motion patterns.}
    \label{fig:L_dynamic_env}
\end{figure}
\subsubsection{Testing Environments:} 
We designed five distinct testing environments, labeled Env1 through Env5, with progressively increasing difficulty based on obstacle proximity to evaluate the performance of the proposed method on static obstacle avoidance tasks, as illustrated in \Cref{fig:static_env}. In Env1, obstacles are spaced 3 meters apart, whereas in Env5, the distance between obstacles is reduced to 1 meter. This gradual reduction in spacing allows for a comprehensive assessment of POVNav's performance across varying levels of obstacle density.

To validate the method on dynamic obstacle avoidance tasks, we designed a corridor environment with dimensions of 30 meters by 6 meters, populated with pedestrians, as illustrated in \Cref{fig:dynamic_env}. We refer to this environment as Env6. The pedestrians' movements followed the social force model proposed by \cite{helbing1995social}, simulating realistic human-like navigation behavior. In all experiments, the robot's maximum linear speed was set to 2 m/s, and its maximum angular speed was constrained to $\pi$ rad/s. The pedestrians moved with varying speeds, with a maximum speed of 1.7 m/s, adding dynamic complexity to the environment. This setup allowed for a comprehensive evaluation of the proposed method's ability to avoid dynamic obstacles.

To increase difficulty for navigation,
we also designed a more complex simulated scenario featuring an L-shaped corridor, as illustrated in \Cref{fig:L_dynamic_env}. The environment includes 10 to 40 pedestrians to emulate realistic navigation conditions with bidirectional human traffic. Specifically, half of the pedestrians are initialized near the robot’s starting position and move toward the goal, while the remaining half begin near the goal and move toward the start. This configuration ensures that the robot continuously encounters pedestrians approaching from both directions, thereby increasing interaction complexity. Furthermore, pedestrian trajectories are intentionally directed to pass near the corner of the L-shaped corridor. This design choice creates partial occlusion in the robot's field of view, mimicking real-world blind-spot scenarios and allowing us to assess the system’s responsiveness to suddenly appearing obstacles.

\begin{figure*}[tbp]
    \centering
    \subfigure[Success rate vs Environments]{
        \includegraphics[height=1.4in, width=2.15in]{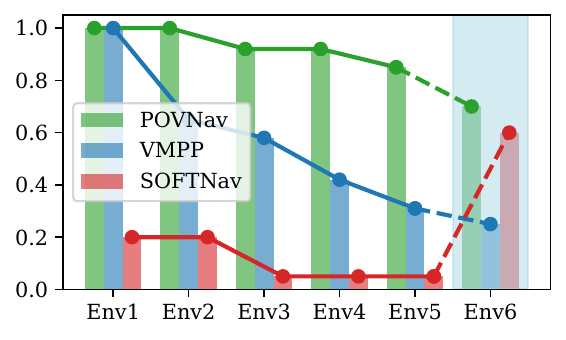}
    \label{fig:success_rate}}
    \subfigure[Path length in Sparse Environment]{
        \includegraphics[height=1.4in, width=2.15in]{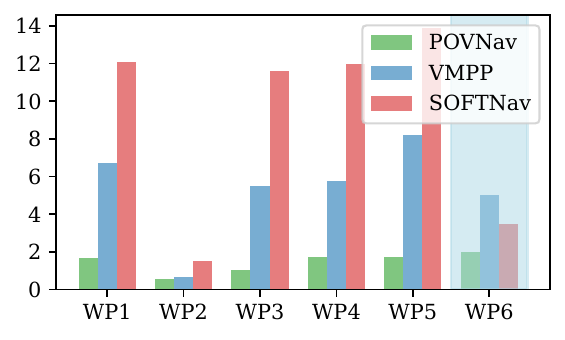}
    \label{fig:path_length_sparse}}
    \subfigure[Path length in Dense Environment]{
        \includegraphics[height=1.4in, width=2.15in]{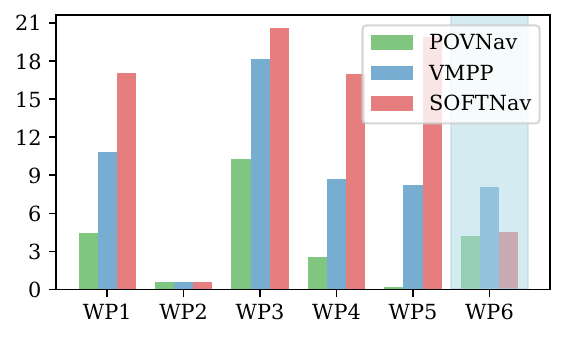}
    \label{fig:path_length_dense}}
    \caption{Performance analysis of the proposed POVNav method compared to the baselines. (a) The $x$-axis represents different testing environments, while the $y$-axis shows the success rate, illustrating the comparative performance of POVNav across various environments. Solid lines depict the trend in success rate across all static environments, and the dashed line indicates the transition from static to dynamic obstacle environment. In (b) and (c), the $x$-axis corresponds to different navigation tasks, where WP stands for Waypoint, and the $y$-axis represents the path length differences from shortest path to goal in meters, demonstrating the efficiency of the proposed method compared to the baselines.}
    \label{fig:comparision_results}
\end{figure*}

\begin{figure}[tbp]
    \centering
    \subfigure[POVNav vs. Reactive Planners]{
        \includegraphics[height=1.2in, width=1.535in]{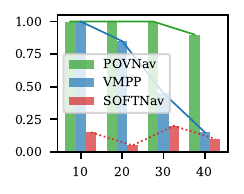}
        \label{fig:L_dynamic_env1}}
    \subfigure[POVNav vs. Modular Planner]{
        \includegraphics[height=1.2in, width=1.535in]{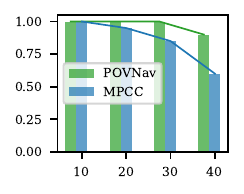}
        \label{fig:L_dynamic_env2}}
    \caption{Performance evaluation of the proposed POVNav method in an L-shaped corridor with varying pedestrian densities. The $x$-axis indicates the number of pedestrians in the scene, while the $y$-axis denotes the success rate. 
    (a) Comparison of POVNav with two reactive baselines: VMPP and SOFTNav, highlighting its superior adaptability under increasing dynamic complexity. The dashed line indicates inconclusive or unstable performance trends. 
    (b) Evaluation against a representative modular baseline, MPCC, demonstrating POVNav's robustness in dynamic environments where the goal is not in front of the robot.}
    \label{fig:L_dynamic_env_comparision_results}
\end{figure}

\subsubsection{Results:}
We start the basic performance evaluation in simulation with relatively simple environments. We first compare the proposed POVNav method with VMPP and SOFTNav across these testing environments. In static environments, 20 randomly generated start and goal points were selected, while in dynamic environments that contains 10 pedestrians, the same task was repeated 20 times to account for the variability introduced by moving obstacles. The results, presented in \Cref{fig:success_rate}, demonstrate that the proposed method consistently outperforms the baselines in all environments. The difference in performance between POVNav and VMPP becomes increasingly evident in more cluttered environments. The increasing gap is due to VMPP's reliance on depth data for trajectory validation, limiting its obstacle detection to the depth sensor's shorter range, whereas the visual horizon cues used by POVNav extend farther. {\em This enables POVNav to take preventive actions for collision avoidance well before VMPP detects obstacles. } Additionally, VMPP requires evaluating more motion primitives to navigate dense obstacles, making the set of primitives sufficient for simpler environments (Env1) inadequate in more complex scenarios. 

The SOFTNav method, while showing lower success rates in static environments due to insufficient feature detection in these scenarios and its lack of goal-directed navigation capabilities, exhibits better performance than VMPP in dynamic environments. This improvement underscores the effectiveness of the optical flow and time-to-transit concept in avoiding moving objects, originally inspired by studies of insect behavior. 

In addition to evaluating the success rate, we also assess the path length. We randomly selected five paths for evaluation in a sparse environment where the gap between obstacles is 2 meters, denoted as WP1 to WP5. Among these, WP2 turned out to be obstacle-free. These paths include the longest diagonal goal, an obstacle-free goal, and intermediate point goals. Additionally, we include a path WP6 in the dynamic environment with three obstacles. Together, WP1 to WP6 represent the navigation tasks in the sparse environment. Given the homogeneous nature of the environment, these paths are sufficient for comparative studies.
We used the same waypoints in a dense static environment where obstacles are spaced 1 meter apart, and 10 pedestrians are included in the corridor navigation task instead of three and the start and goal points remain the same, as shown in \Cref{fig:dynamic_env}, representing navigation in the dense environment. 
We performed 10 successful trials for each path. The path length was recorded during the trials, and the deviation from the shortest obstacle-free path for each method is shown in \Cref{fig:path_length_sparse} and \Cref{fig:path_length_dense}.
Our comparative analysis demonstrates that POVNav consistently follows a shorter path than both VMPP and SOFTNav, except in WP2 across both environments. In the sparse environment, SOFTNav deviated slightly in WP2 to maintain equal distance from sorounding obstacles detected via optical flow, while POVNav and VMPP followed the shortest path due to the absence of obstacles on the way and they did not get affected by the surrounded environment. Generally, SOFTNav tended to follow longer paths, as its goal-direction-agnostic optic flow detections led to unnecessary deviations.
The performance of VMPP is highly dependent on the spacing between candidate motion primitives, as it involves converting the continuous velocity space of the dynamic window into discrete options. This method requires careful tuning of control parameters for optimal results. In densely cluttered environments, a large number of trajectories need to be evaluated; however, we used the same parameters across all methods that was optimized for the sparse environment. VMPP's performance did not generalize well across different environments, as parameters optimized for Env1 did not yield similar results elsewhere.
In contrast, POVNav exhibited less sensitivity to changes in the environment, suggesting its adaptability across various conditions without significant parameter adjustments. Interestingly, SOFTNav's performance improved significantly in the dynamic dense environment, with its path length becoming comparable to that of the proposed method. This environment is ideal for SOFTNav, as it continuously receives features from nearby obstacles, allowing it to find the correct control inputs. This aligns with the insect-inspired navigation behavior that motivated the development of SOFTNav.

Experiments in the straight corridor provided valuable insights into the overall performance of POVNav in comparison to VMPP and SOFTNav. However, in this setup the goal was always aligned with the robot's heading direction. To further assess the robustness of each method, we extended with a longer L-shaped corridor environment in which the goal is not directly visible from the start, as illustrated in~\Cref{fig:L_dynamic_env}. This environment demands not only reactive capabilities but also effective decision-making under partial observability and dynamic occlusion.
We referr to these environments as \textit{DyEnv1} (10 pedestrians) through \textit{DyEnv4} (40 pedestrians). 
The robot starts at one end of the corridor, and the goal is placed at the far end, approximately 30 meters forward and 20 meters to the right. Each method was evaluated over 20 independent trials, and the performance results are summarized in ~\Cref{fig:L_dynamic_env_comparision_results}.

For a fair comparison, we tuned the parameters of the four methods (POVNav, VMPP, SOFTNav and MPCC) to perform optimally in the lowest-density setting (\textit{DyEnv1}) and kept these parameters fixed across the higher-density scenarios. As shown in~\Cref{fig:L_dynamic_env1}, under these conditions, both POVNav and VMPP achieved good performance in \textit{DyEnv1}. 
However, in this L-shaped  environment, we started observing that SOFTNav struggled consistently due to its strong dependence on optical flow features, which are less reliable in low-texture, sparsely colored simulated environments. Additionally, the L-shaped geometry posed a significant challenge for SOFTNav, as it frequently misinterpreted corner features and became stuck at the turn.

As pedestrian density increased, VMPP’s performance deteriorated significantly, as evidenced by its decreasing success rate. This degradation is primarily due to increased reactive latency stemming from the growing number of motion primitives, which were originally tuned for optimal performance in the low-density scenario (\textit{DyEnv1}). In more crowded environments, the method requires denser and more frequent evaluation of motion primitives to ensure safe responses, which in turn increases computational load and reduces responsiveness. In contrast, POVNav maintains its responsiveness regardless of the number of pedestrians, as it depends solely on the visual horizon and image-space features for decision-making. This lightweight design enables faster and more proactive reactions to dynamic obstacles. We capped the pedestrian count at 40, beyond which the scene became excessively congested. At this point, the performance disparity between POVNav and the other baselines became substantially more evident.

POVNav consistently outperformed VMPP and SOFTNav due to its early and efficient decision-making strategy. Its computational simplicity allows for rapid evaluation of feasible paths and proactive maneuvers before entering congested regions. Importantly, as pedestrians enter the robot's field of view, they dynamically reshape the visual horizon, thereby modifying the POG and influencing the visual path. When a pedestrian approaches closely, the proximity feature triggers an immediate retreat action. Remarkably, the same set of parameters tuned in (\textit{DyEnv1}) remained effective across all subsequent scenarios, indicating the robustness and generalizability of the proposed approach under increasing dynamic complexity. 

We further evaluated the robustness of POVNav in dynamic environments by comparing it against MPCC. Note however, different from VMPP and SOFTNav, the MPCC approach relies on following a precomputed global path that avoids static obstacles. For a fair comparison, we designed this global path such that the MPCC successfully reaches the goal in the least crowded setting (\textit{DyEnv1}), and used the same path across all other experiments to maintain consistency. MPCC outperforms reactive baselines like VMPP and SOFTNav in low-density settings, primarily due to its reliance on the global path, which inherently steers the robot away from walls and other static structures. However, as pedestrian density increases, the optimization overhead in MPCC becomes more pronounced. The need to replan more frequently in cluttered scenes leads to increased computational delays and, consequently, a noticeable decline in success rate, as shown in \Cref{fig:L_dynamic_env2}.

Lastly, we analyzed the advantages of POVNav over a learning-based approach by comparing it with iPlanner, a recently proposed method with publicly available training and evaluation code. We first trained iPlanner in a simulated forest environment and evaluated both iPlanner and POVNav in this training environment, as well as in a previously unseen indoor office environment to assess generalization. Next, we incorporated the office environment into the training dataset and retrained iPlanner. Both methods were then tested again in both environments to evaluate improvements in adaptability. To adapt to iPlaner and make the comparison fair, both methods used depth image for perception. 

For each environment, we randomly sampled 20 goal locations for evaluation. In the office environment, we excluded goal configurations where U-shaped obstacles entirely blocked the path between the start and goal, as neither method (POVNav being purely local planners) could handle such topologically infeasible cases. The performance results from these experiments are summarized in Table~\ref{tab:planner_comparison}.

\begin{table}[tbp]
    \centering
    \caption{Comparison of POVNav and iPlanner in terms of robustness to environmental changes. \textbf{Scenario 1:} iPlanner is trained only in a simulated forest environment and tested in both the forest and an unseen office environment. \textbf{Scenario 2:} iPlanner is trained in both forest and office environments and evaluated in the same environments. Success rates are reported for each case.}
    \label{tab:planner_comparison}
    \resizebox{\linewidth}{!}{%
    \begin{tabular}{lcc}
        \toprule
        \textbf{iPlanner (Success Rate)} & \textbf{POVNav (Success Rate)} \\
        \midrule
        \textit{(Forest-trained only)} & \\
        Forest: 95\% \quad Office: 15\% & Forest: 100\% \quad Office: 85\% \\
        \textit{(Forest + Office-trained)} & \\
        Forest: 100\% \quad Office: 90\% & Forest: 100\% \quad Office: 90\% \\
        \bottomrule
    \end{tabular}
    }
\end{table}

\begin{figure*}[tbp]
  \centering
  \subfigure[Map]
  	{\label{fig:ablation_a}\includegraphics[width=4.25cm]{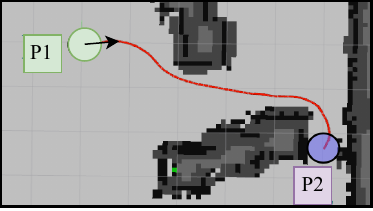}}
  \subfigure[POG and HOG]
  	{\label{fig:ablation_b}\includegraphics[width=4.25cm]{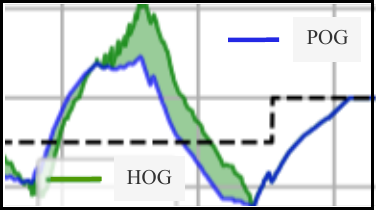}}
  \subfigure[POG, HOG and $\phi$]
  	{\label{fig:ablation_c}\includegraphics[width=4.25cm]{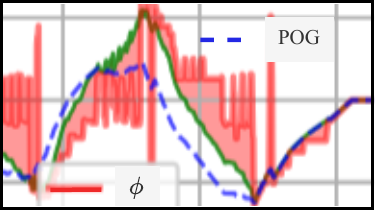}}
  \subfigure[Angular Control]
  	{\label{fig:ablation_d}\includegraphics[width=4.25cm]{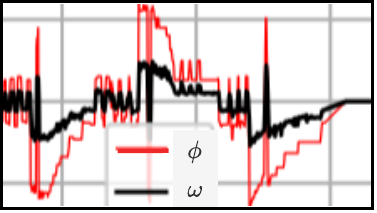}}
  \caption{\small Shows the role of different components in POVNav. The $x$-axis in all the subplots represents time and $y$-axis represents error value. (a) Dashed line shows the obstacle detection status (non-zero value means obstacle detected). (b) Shows the effect of visual path planning on obstacle avoidance. (c) Shows $\phi$ controls $\omega$. (d) Shows the effect of $\lambda$ on $v$.\vspace{-12pt}}\label{fig:tracking}  
\end{figure*}

\begin{figure}[tbp]
    \centering        
    \includegraphics[width=5.5cm, height=3.4cm]{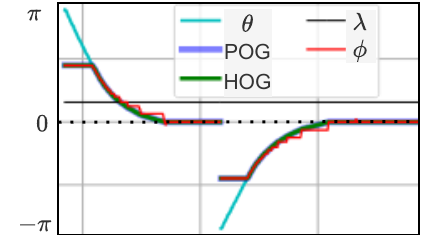}
    \caption{\small Illustration of obstacle-free goal navigation with POVNav ($\lambda = \lambda_0$) for two goal directions ($\pi$ to $0$, $-\pi$ to $0$). The discontinuity in the plot represents the change in navigation goal. $\theta$ is the heading angle of the robot.\vspace{-5pt}}\label{fig:A1} 
\end{figure}

It is clearly evident from Table~\ref{tab:planner_comparison} that the learning-based planner, iPlanner, exhibits a significant drop in performance when tested in an unseen environment especially if there is a domain shift. Specifically, when trained solely in the simulated forest environment, iPlanner achieves a high success rate of 95\% in the forest, but its performance deteriorates sharply to 15\% when deployed in the office environment. This highlights a key limitation of end-to-end learning-based methods, their sensitivity to distributional shifts and lack of generalization when confronted with novel or structurally different environments.
In contrast, POVNav maintains strong performance across both training and testing environments, achieving 100\% success in the forest and 85\% in the office setting. 

Furthermore, when iPlanner is retrained with data from both forest and office environments, its performance improves to 90\% in the office, matching the performance of POVNav. However, this improvement comes at the cost of increased data collection and retraining effort. POVNav, on the other hand, achieves similar performance without requiring any additional fine tunning of parameters, allowing the system to dynamically adjust its behavior based on environmental conditions or task-specific requirements. 
The modular and interpretable nature of POVNav offers a compelling trade-off between adaptability and transparency, especially in safety-critical applications.

We also evaluated the computational time of POVNav and the baseline methods. POVNav demonstrated superior efficiency, with an average of 0.04 seconds per action, compared to 0.22 seconds for VMPP and 0.14 seconds for SOFTNav. This improvement stems from POVNav's image-based planning, which eliminates the need for evaluating multiple 3-D trajectories like VMPP, and is faster than SOFTNav's optical flow and time-to-transit computations, {\em making it ideal for real-time applications.}

\begin{table}[tbp]
    \centering
    \caption{Success Rate Comparison in Ablation Studies. 1.Obstacle-free Navigation, 2.Static Obstacle Avoidance and 3. Dynamic Obstacle Avoidance.}
    \label{tab:ablation_success_rate}
    \begin{tabular}{rrrr}
        \toprule
        \textbf{Scenario} & \textbf{POG Only} & \textbf{POG+HOG} & \textbf{Full System} \\
        \midrule
        1 & 100\% & 100\% & 100\% \\
        2 & 0\% & 85\% & 100\% \\
        3 & 0\% & 45\% & 80\% \\
        \bottomrule
    \end{tabular}
\end{table}

\begin{table}[tbp]
    \centering
    \caption{Path Length Comparison in Ablation Studies. All waypoints are 32 meters apart in obstacle-free, static, and dynamic environments. The obstacles are 2 meters apart in static environment and the dynamic environment includes 6 pedestrians.}
    \label{tab:ablation_path_length}
    \begin{tabular}{rrrr}
        \toprule
        \textbf{Scenario} & \textbf{POG Only} & \textbf{POG+HOG} & \textbf{Full System} \\
        \midrule
        1 & 32m & 32m  & 32m \\
        2 & N/A (Fail) & 37.16m & 38.95m \\
        3 & N/A (Fail) & 34.21m & 36.21m \\
        \bottomrule
    \end{tabular}
\end{table}

\subsection{Ablation Studies}
We conduct ablation studies across three scenarios in a simulated environment to evaluate the impact of different components in our visual navigation framework: (i) obstacle-free navigation, (ii) navigation while avoiding static obstacles, and (iii) navigation while avoiding dynamic obstacles. Each experiment is designed to isolate and test the contribution of specific features:

\begin{itemize}
    \item \textbf{POG}: We assess how the absence of HOG and visual path affects the robot's performance in navigating complex or cluttered areas.
    \item \textbf{POG+HOG}: The importance of HOG for robust visual navigation is evaluated, focusing on the success rate during static and dynamic obstacle avoidance tasks.
    \item \textbf{Visual Path}: We compare navigation performance with and without the visual path planning component.
\end{itemize}

A comparison of success rates across different navigation scenarios, highlighting the distinct roles of POG, HOG, and the visual path in ensuring successful navigation is shown in \Cref{tab:ablation_success_rate}. In an obstacle-free environment, all methods successfully reached the goal, indicating that POG alone can effectively map the global goal onto the image boundary and guide the robot without obstacles. However, in environments with obstacles, the {\em POG-Only} method had a $0\%$ success rate, as it lacked the capability to avoid obstacles. This suggests that, while POG alone can be sufficient for guiding the robot in open spaces, it requires additional support for safe navigation in cluttered environments.
Table~\ref{tab:ablation_path_length} shows that all methods followed the same path in the absence of obstacles, indicating that HOG and the visual path are not required in obstacle-free navigation tasks. When HOG was combined with POG, the success rate increased to $85\%$ in static and $45\%$ in dynamic environments, demonstrating that HOG provides a local goal perspective that optimizes exploration and navigation objectives and help avoiding obstacles by deviating from POG. This supports the hypothesis that HOG aids in navigating complex environments by helping to avoid nearby obstacles.
Adding the visual path further improved the success rate, achieving $100\%$ in static and $80\%$ in dynamic environments. This improvement is attributed to the visual path's ability to add a layer of safety by maintaining a safe distance from obstacles, even if this requires deviating from the local goal. However, this also led to an increase in path length, suggesting a trade-off between path efficiency and safety in more challenging navigation tasks.

\begin{figure*}[tbp]
  \centering
  \subfigure[Real World: POG and HOG]
  	{\label{fig:real_ablation_a}\includegraphics[width=0.49\linewidth]{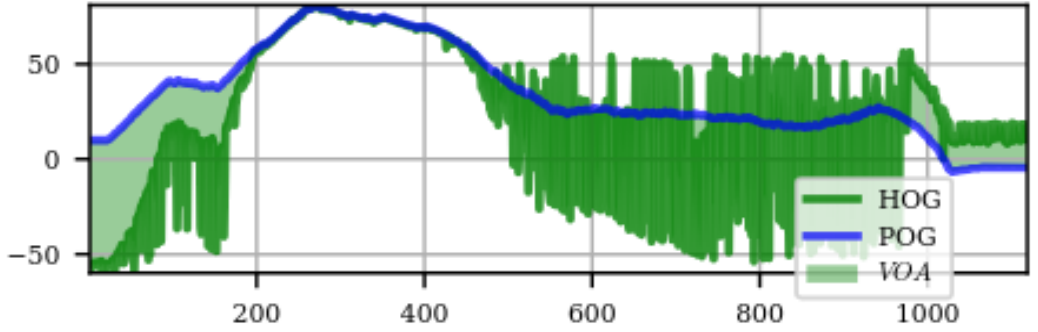}}
  \subfigure[Real World: HOG and Visual Path]
  	{\label{fig:real_ablation_b}\includegraphics[width=0.49\linewidth]{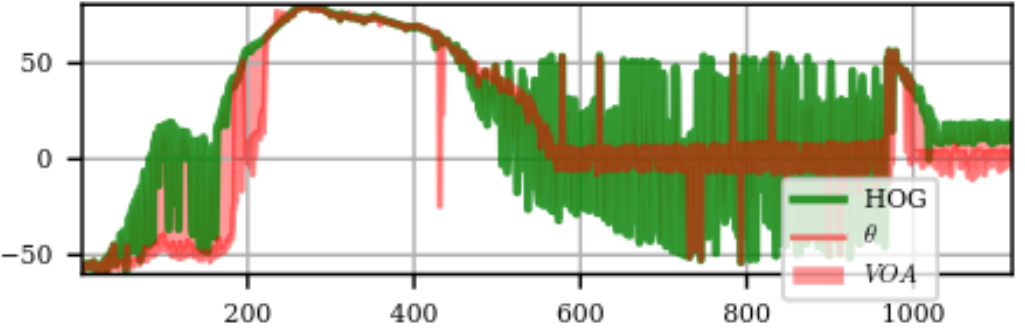}}
  \caption{\small Real-world navigation task. (a) Sub-goal selection with noisy segmentation, illustrating the effect of segmentation noise on sub-goal stability. (b) Obstacle avoidance, where frequent fluctuations in HOG highlight segmentation noise. The Visual Obstacle Avoidance (VOA) regions, indicated between HOG and POG as well as between POG and alignment features, reflect the system's adaptive response to maintain navigability despite noise.}
\label{fig:real_ablation}  
\end{figure*}

\begin{figure*}[tbp]
    \centering        
    \includegraphics[width=17cm]{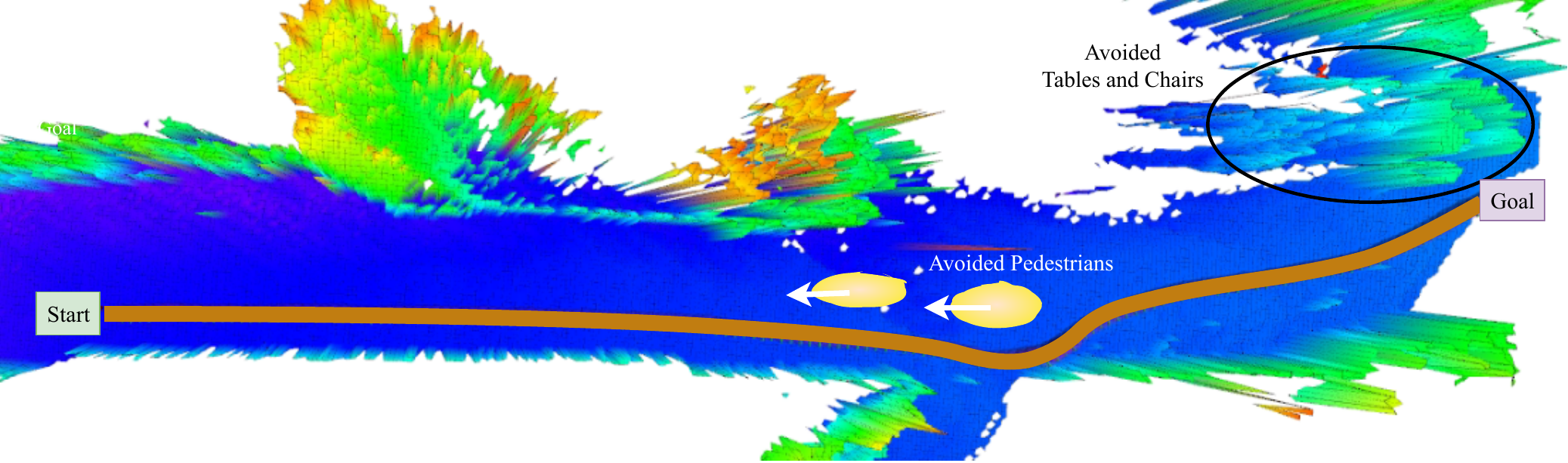}
    \caption{Illustration of Jackal robot in the real-world navigating around static obstacles such as walls, tables, and chairs, as well as dynamic obstacles like pedestrians. The elevation map is used solely for visualizing the path followed and the obstacles encountered.}\label{fig:obs_avoid_real}
\end{figure*}

\begin{figure*}[tbp]
  \centering
  \subfigure[$t_1$]
  	{\label{fig:t1}\includegraphics[width=4.25cm]{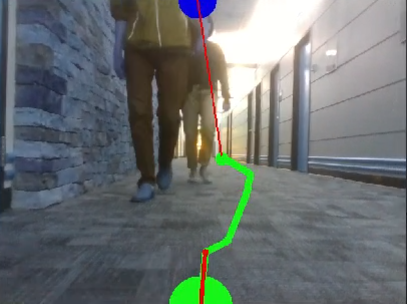}}
  \subfigure[$t_2$]
  	{\label{fig:t2}\includegraphics[width=4.25cm]{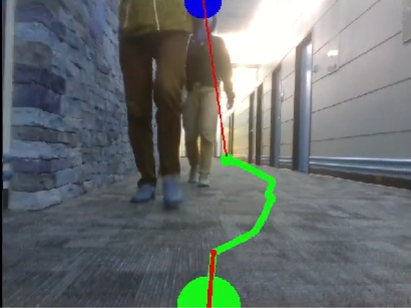}}
  \subfigure[$t_3$]
  	{\label{fig:t3}\includegraphics[width=4.25cm]{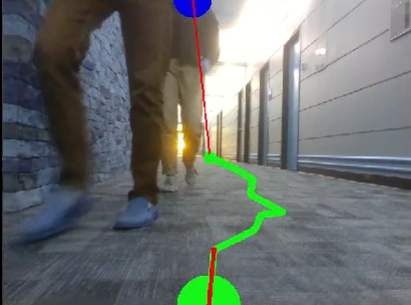}}
  \subfigure[$t_4$]
  	{\label{fig:t4}\includegraphics[width=4.25cm]{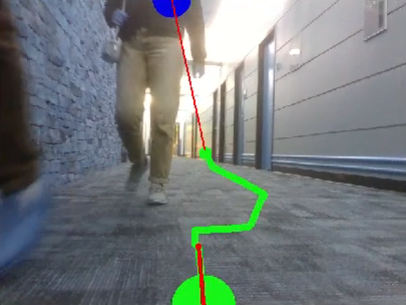}}
  \caption{\small Shows the POVNav planning image from timestamp $t_1$ to $t_4$ during human pedestrians avoidance. From (a) to (d), the change in visual path (shown in green color) in real-time with changing obstacle positions shows the active visual path planning of POVNav to avoid dynamic obstacles in real-world.}
\label{fig:real_active_path}  
\end{figure*}

By analyzing these cases, we demonstrate that each component plays a crucial role in enhancing the success rate of navigation, particularly in scenarios involving static and dynamic obstacle avoidance. To gain further insight into the framework, we recorded and analyzed the behavior of all visual features during both obstacle-free and obstacle-rich navigation tasks.

To see the role of POG, we performed an experiment in which we define different goals in random directions of an obstacle-free, simulated environment. The results are shown in \Cref{fig:A1} where the robot aligns itself towards the goal direction and then it maintains the heading angle until it reaches the goal. Note that $\theta$, POG, HOG, $\phi$ overlaps except for two intervals ($\pi~to~\pi/2, -\pi~to~-\pi/2$) as a result of the non-uniform mapping of the goal direction on the image border. This shows that the implemented mapping function maintains the correlation of POG with $\theta$. In our case, the POG updates itself with the same value during that interval. This also shows that the POG itself is able to navigate the robot in an obstacle-free environment and can be considered as a global guide in POVNav.

\begin{figure*}[tbp]
  \centering
  \subfigure[On-road to Off-road Transition]
  	{\label{fig:a2b}\includegraphics[width=8.6cm]{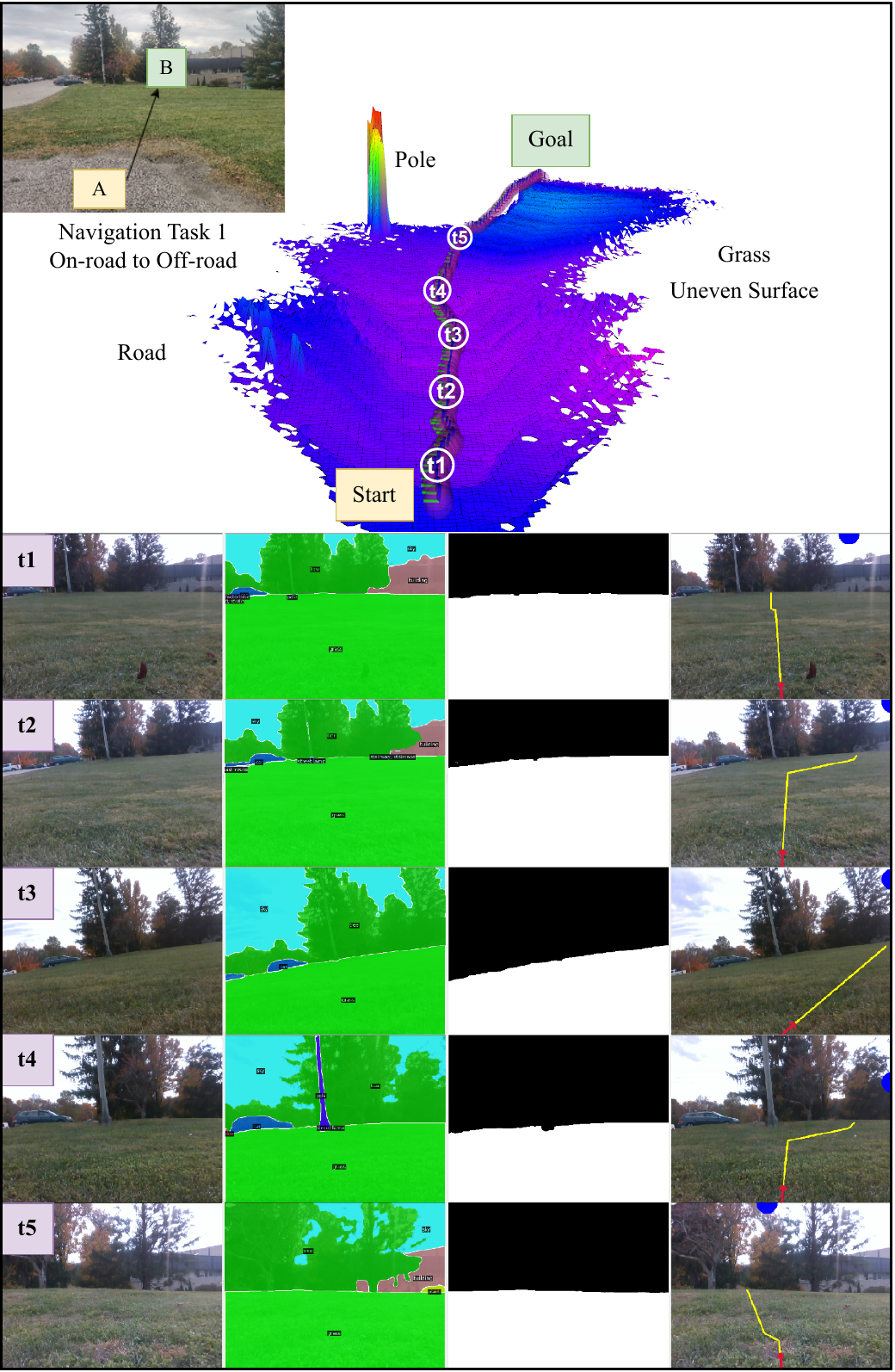}}
   \subfigure[Avoiding Sparse Obstacles]
  	{\label{fig:c2d}\includegraphics[width=8.6cm]{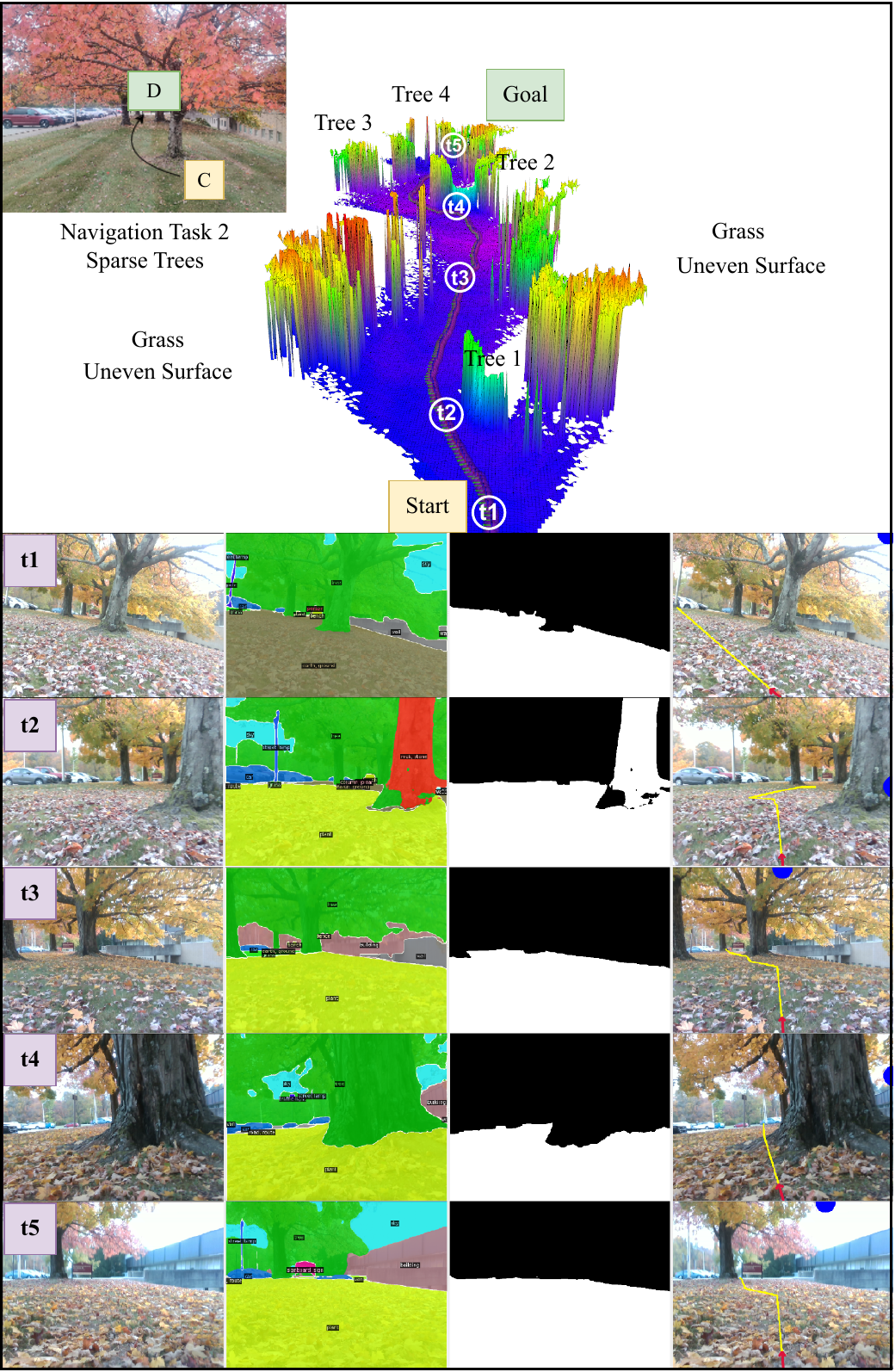}}
  \caption{\small Visual representation of two distinct navigation tasks in real-world environments. In the top left, a third-person view displays the environment with marked start and goal points. A 2.5D elevation map is included to provide context on the terrain and the robot's navigation path. Along the path, five key points are selected, each representing the robot's state at different timestamps. For each state, four images are presented sequentially from left to right: the RGB image perceived by the robot, the segmented image generated by Mask2Former~\citep{cheng2022masked}, the binary segmented image defining navigability, and the path planned by the robot.}
\label{fig:a2d}  
\end{figure*}

We introduce obstacles into the environment to evaluate how POG, HOG, and Visual Path Planning collaboratively guide the robot to its goal while avoiding obstacles. \Cref{fig:ablation_a} illustrates the environment setup, with specified start and goal positions, as well as visible obstacle locations. \Cref{fig:ablation_b} highlights the sub-goal selection process, which effectively balances the trade-off (indicated by the green region) between two conflicting objectives, guiding the robot towards the goal. The alignment of HOG with POG enhances traversability by maximizing the robot's ability to navigate through complex spaces. Once a sub-goal is chosen, the robot interacts dynamically with the environment, using visual path planning to actively avoid obstacles, as shown in \Cref{fig:ablation_c}. Finally, \Cref{fig:ablation_d} presents the calculated control inputs derived from alignment features, providing a clearer view of the framework's complete operation.

\subsection{Robustness in the Real-world} 
We conducted the following experiments to evaluate the robustness of the proposed method in real-world settings:
\begin{itemize}
    \item Navigation with noisy segmentation methods.
    \item Point-to-point navigation in different scenarios.
    \item Navigation in dense forest trails.
    \item Navigation under extreme weather conditions.
    \item Kilometer-scale navigation leveraging visual-inertial odometry, with GPT-3.5 utilized as a navigability parser.
\end{itemize}

\subsubsection{Navigation with Noisy Segmentation Techniques}
Real-world image segmentation methods often exhibit inconsistencies in predicting precise boundaries between segments. This challenge is observed across various segmentation approaches, including color-based, depth-based, surface normal-based, and deep learning-based methods, all of which introduce noise in boundary pixel prediction in the real-world. These inaccuracies impact the exploration objective of POVNav, leading to significant fluctuation in the HOG. Nevertheless, the proposed framework's active path-planning approach effectively mitigates these fluctuations.

As shown in \Cref{fig:real_ablation_a}, the POG path remains relatively smooth, while HOG exhibits noticeable instability due to segmentation noise. The POVNav framework mitigates these fluctuations by planning a visual path but only executing the initial action. This selective approach reduces the impact of noise on HOG calculations, as illustrated in \Cref{fig:real_ablation_b}. Even with noisy segmentation, \Cref{fig:obs_avoid_real} demonstrates the Jackal robot's ability to navigate between waypoints and avoid obstacles. Snapshots at four distinct timestamps in \Cref{fig:real_active_path} reveal the robot's pedestrian avoidance behavior as the path dynamically updates.

Despite segmentation noise, the robot consistently reaches its goal in most trials, highlighting POVNav's robustness in noisy environments.

\begin{figure*}[tbp]
  \centering
  \subfigure[Selective Navigation Behavior]
  	{\label{fig:e2f}\includegraphics[width=8.6cm]{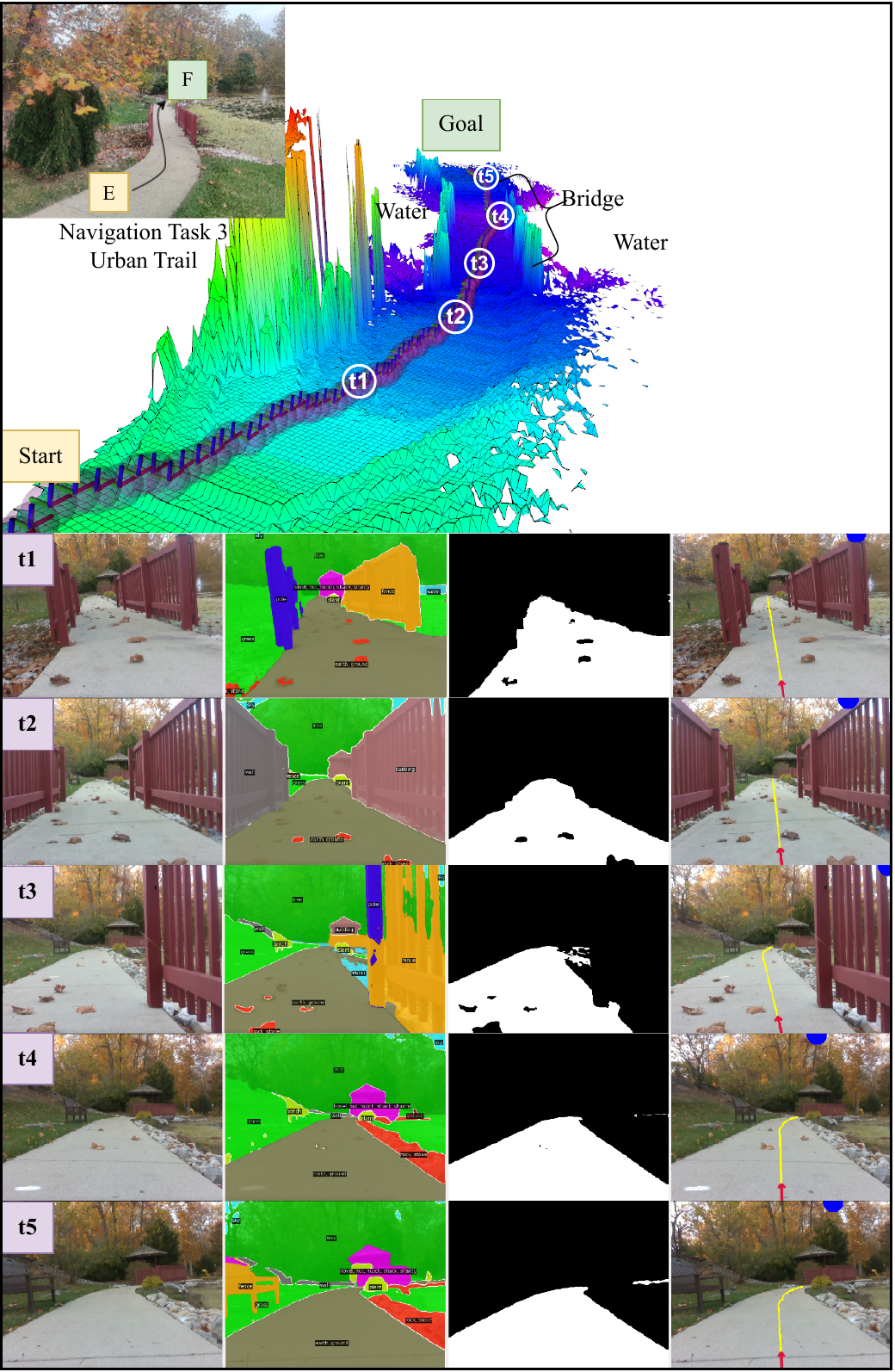}}
   \subfigure[Navigating through Dense Forest Trail]
  	{\label{fig:g2h}\includegraphics[width=8.6cm]{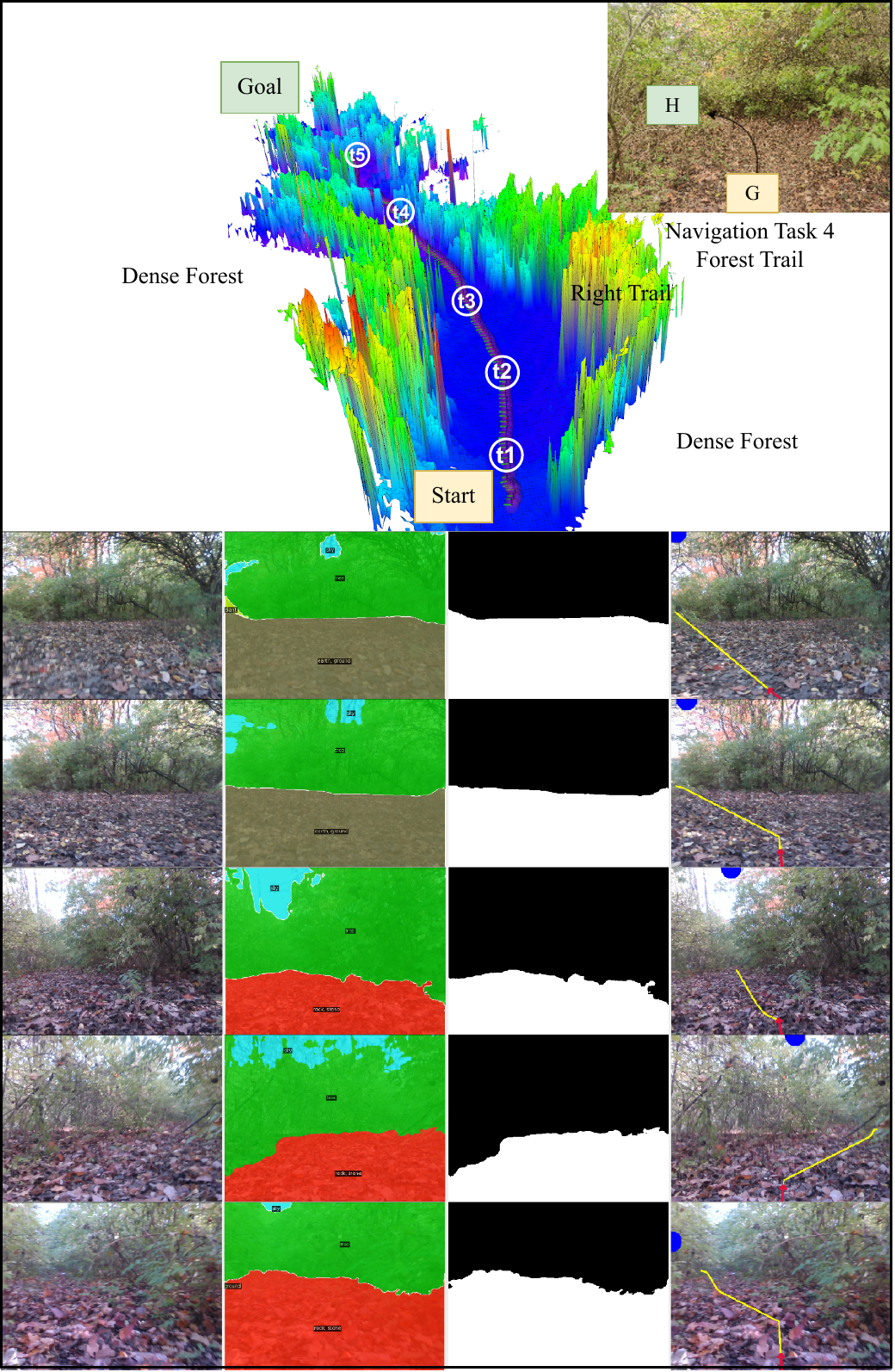}}
  \caption{\small Visual representation of two distinct navigation tasks in real-world environments. In the top left in (a) and top right in (b), a third-person view displays the environment with marked start and goal points. A 2.5D elevation map is included to provide context on the terrain and the robot's navigation path. Along the path, five key points are selected, each representing the robot's state at different timestamps. For each state, four images are presented sequentially from left to right: the RGB image perceived by the robot, the segmented image generated by Mask2Former~\citep{cheng2022masked}, the binary segmented image defining navigability, and the path planned by the robot.}
\label{fig:e2h}  
\end{figure*}

\subsubsection{Point-to-Point Navigation in Different Scenarios}
We conducted rigorous real-world navigation experiments in four distinct scenarios to further test the navigation robustness: 
\begin{itemize}
    \item \textbf{A to B}: Transition from on-road to off-road conditions.
    \item \textbf{C to D}: Navigation through sparse obstacles.
    \item \textbf{E to F}: Selective navigation where the robot must avoid grass while heading towards the goal.
    \item \textbf{G to H}: Navigation through dense forest terrain.
\end{itemize}
Each experiment was conducted five times under real-world conditions to evaluate the proposed method's consistency and reliability across diverse environments. Performance was measured through three key metrics: success rate, the number of human interventions required per run, and the frequency of redefined navigability criteria. Table~\ref{tab:point_to_point_results} summarizes the results, showing stable performance and adaptability of the proposed method across varied navigation challenges. As no changes in navigability criteria were observed across the five runs in each environment, this measure is not included in the table.

\begin{figure*}[tbp] 
    \centering
    \includegraphics[width=\linewidth]{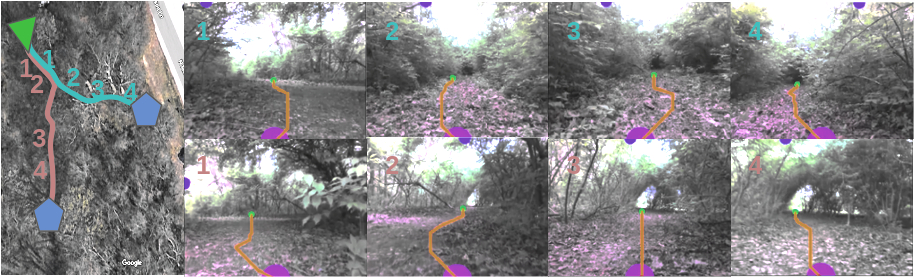}
    \caption{Illustration of Navigation task in unstructured environments. The green marker indicates the robot's starting position, while the blue markers denote the goal positions. The purple marker represents the robot's state in the image space, and the orange lines depict the planned safe paths.}
    \label{Fig:rw_exp_forest_trail1}
\end{figure*}

\begin{figure*}[tbp]
     \centering
     \subfigure[Road $=$ Navigable; Snow $\neq$ Navigable]{\includegraphics[width=0.33\linewidth]{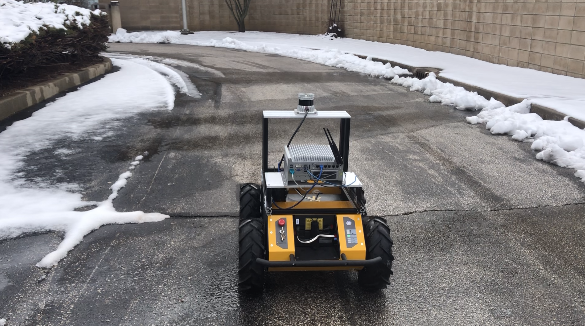}\label{fig:out_exp1}}
     \subfigure[Road $\neq$ Navigable; Snow $=$ Navigable]{\includegraphics[width=0.33\linewidth]{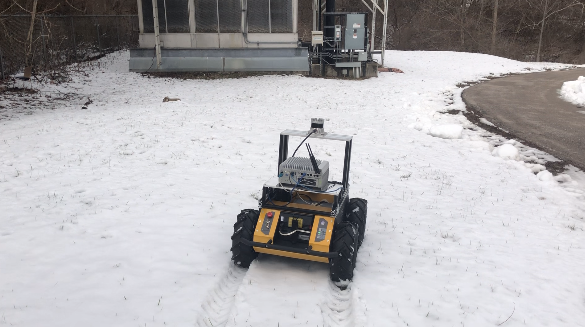}\label{fig:out_exp2}}
     \subfigure[Road $=$ Navigable; Snow $=$ Navigable]{\includegraphics[width=0.33\linewidth]{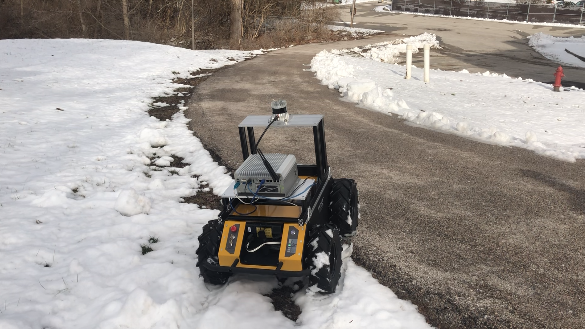}\label{fig:out_exp3}}
    \caption{Navigation under extreme weather conditions. The robot is demonstrating the selective navigable behavior using POVNav. In (a), snow is designated as non-navigable while the road is navigable. In (b), snow is considered navigable and the road is non-navigable. In (c), both snow and road are treated as navigable.}\label{fig:Sel_nav}
\end{figure*}

As shown in Table~\ref{tab:point_to_point_results}, the robot successfully navigated between terrain types (A to B) without interventions. In sparsely cluttered environments (C to D), the system demonstrated consistent performance, requiring only one human intervention due to segmentation errors, underscoring its robustness in low-density scenarios. During selective navigation tests (E to F), the robot achieved a high success rate with no human input, indicating its efficiency in targeted navigation tasks. However, dense forest navigation (G to H) presented greater challenges, resulting in a lower success rate with more frequent interventions. This was primarily due to limitations in the segmentation method~\citep{cheng2022masked}, which struggled to accurately interpret the forest environment's semantics. Repeating this experiment with a domain-adaptive method~\citep{chen2022cali} trained on a small deployment dataset led to a $100\%$ success rate, suggesting that Mask2Former~\citep{cheng2022masked} underperformed compared to domain-adaptive approaches like CALI~\citep{chen2022cali}. 
A representative autonomous navigation run is visually presented in \Cref{fig:a2d} and \Cref{fig:e2h}. We have used both the robots - Husky and Unitree Go1 to accomplish this task.

\begin{table}[tbp]
    \centering
    \caption{Success Rate and Human Intervention Analysis in Different Scenarios}
    \label{tab:point_to_point_results}
    \begin{tabular}{rrrr}
        \toprule
        \textbf{Scenario} & \textbf{Success Rate (\%)} & \textbf{Interventions}\\ 
        \midrule
        A to B  & 100\% & 0 \\
        C to D  & 100\% & 1 \\
        E to F  & 100\% & 0 \\
        G to H  & 80\% & 2 \\
        \bottomrule
    \end{tabular}
\end{table}

\subsubsection{Navigation in Dense Forest Trails}
This experiment aims to validate the performance of POVNav and examine how segmentation methods behave in dense forest environments. Specifically, we assess whether these methods yield consistent results in such settings and evaluate POVNav's capability to handle segmentation inaccuracies. Two trails were selected within a dense forest, as shown in \Cref{Fig:rw_exp_forest_trail1}, with a goal point set 30 meters away. The robot's planned paths at five different waypoints are captured in snapshots for each trail.
Each trail was tested five times. On Trail 1 (top trail), the robot consistently reached the goal in all trials. However, on Trail 2 (bottom trail), the robot failed to reach the goal in two out of five trials due to insufficient segmentation accuracy on the more challenging terrain. While improving segmentation accuracy is beyond the scope of this study, we anticipate that the proposed method's performance would further improve with advancements in segmentation techniques. 
This experiment highlights the challenges of accurate segmentation in dense forest environments, where existing methods often struggle to identify navigable regions effectively. 
We observed that domain-adaptive methods outperform ``segment anything" techniques trained on large, general datasets in these complex terrains.

\begin{figure*}[tbp] 
    \centering
    \includegraphics[width=\linewidth]{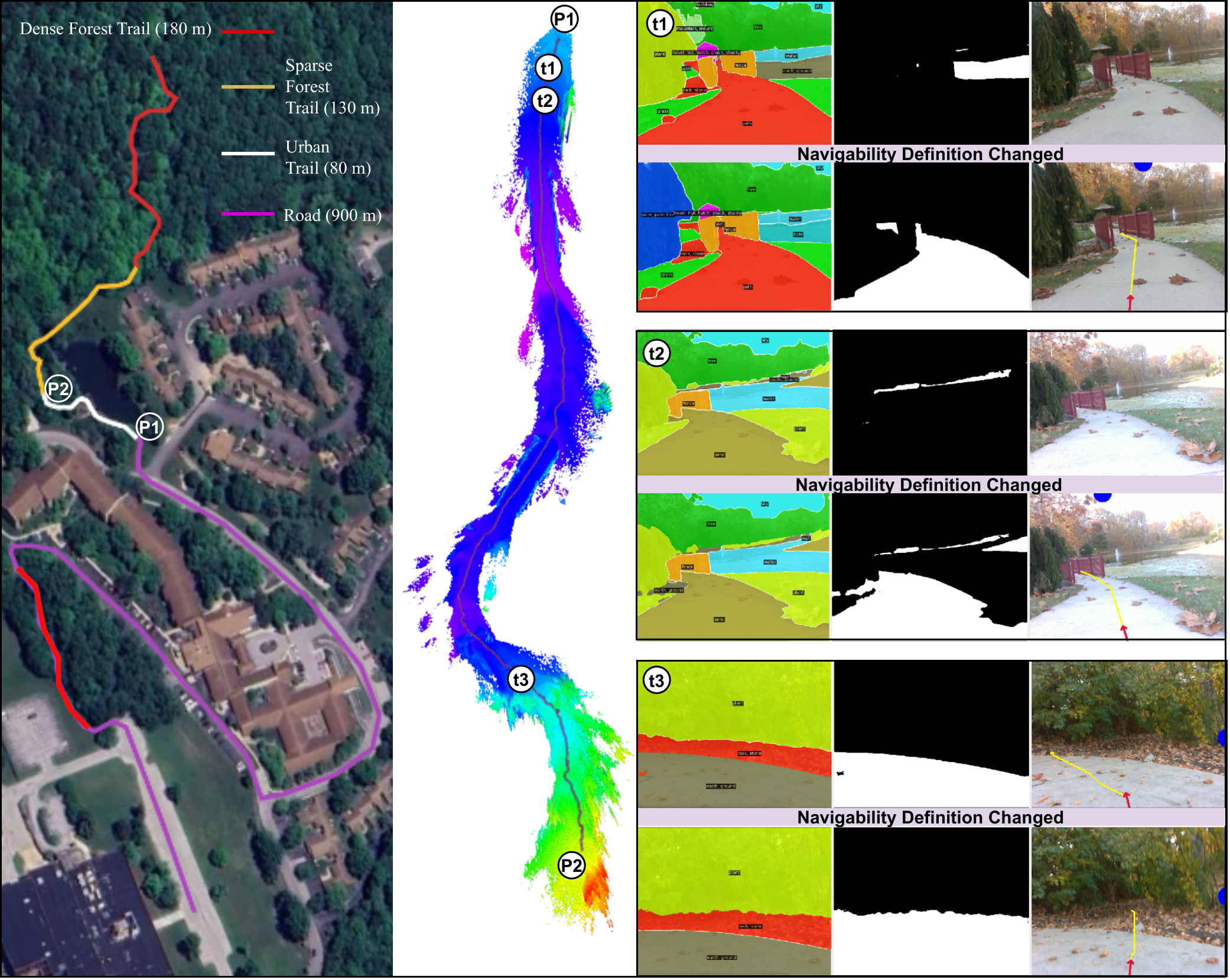}
    \caption{Illustration of KM-Sclae navigation task. On the left, the path travelled by the robot are shown on the map with different colors. From point P1 to P2 the path in while color, we show the elevation map and path travelled by the robot on that map, shoen in the middle. At time t2, t2 and t3, the robot encountered that the naviaglity definitions are not suffient and hence used GPT-API to correct it. The segmented image, image after parsing navigability and the planning image (from left to right) are shown in the right most column. For each timestamp, the images before calling the GPT-API and after are shown.}
    \label{fig:km_scale_nav}
\end{figure*}

\subsubsection{Navigation under Extreme Weather Conditions}
We evaluated the performance of the proposed method in a navigation task following a heavy snowfall. The objective of this experiment was to assess the readiness of POVNav for real-world applications, particularly in situations that demand greater flexibility and adaptability in response to environmental challenges. Since, there are only black and white pixels in the scene, we used color-based segmentation method for this experiment.
During the navigation tasks, we observed sudden drift in the visual odometry; however, this did not compromise the execution of the navigation task. The local observations utilized by POVNav effectively guided the robot, allowing it to handle localization drift without significantly impacting safety, although the accuracy of reaching the goal have been affected. 
POVNav exhibits the capability to selectively navigate over snow or avoid it, depending on the conditions. The modular architecture of POVNav allows for modifications to the navigability definition vector, which is employed to generate the Navigability Image. This flexibility facilitates the demonstration of selective navigation behavior by altering the navigability definitions. Figure~\ref{fig:Sel_nav} illustrates three scenarios in which the robot navigates using distinct definitions of navigability vector. 

\subsubsection{Kilometer-scale Navigation leveraging visual-inertial odometry and GPT-3.5 as a navigability parser}
We conducted a kilometer-scale navigation task where the Husky robot was given a direction to go defined in the local frame and we manipulated the direction clue at difficult places where a local planner typically suffers. This experiment aimed to address two primary questions: (1) How do human-defined navigability definitions align with complex, real-world environments? (2) How can changing navigability definitions be handled, given the potential inaccuracies of segmentation methods?

In our observations, POVNav generally performed well when segmentation models accurately classified segments, and when human-defined navigable classes were adequate, especially in structured environments like urban trails and paved roadways. However, we found that navigability definitions need to be updated when transitioning between different environments, as illustrated by the color-coded paths in \Cref{fig:km_scale_nav}. For example, navigating on a sidewalk requires avoiding grass, whereas in an urban forest environment, a bridge crossing water or mud might be preferable over grassy shortcuts, which can be challenging even for humans to recognize those grass on top of mud or water as non-navigable. In such cases, the shortest path may lead the robot over grass; however, human-like navigation would prefer the bridge.

\begin{figure*}[tbp] 
    \centering
    \includegraphics[width=\linewidth]{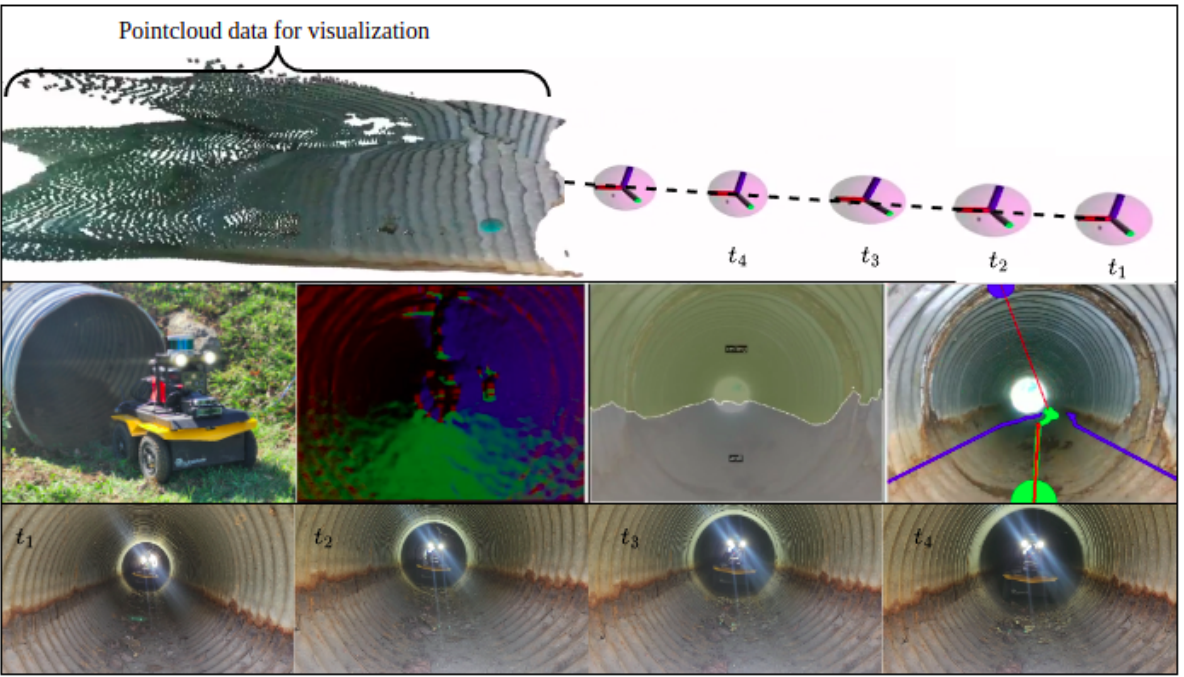}
    \caption{Illustration of the culvert inspection task. The top row shows the robot's pose, its traveled path within the culvert, and the RGB-D point cloud captured at the current timestamp. The middle row begins with an image of the experimental setup, where the culvert has a 30-inch radius. The modified robot, equipped with two LED lights for enhanced visibility, was outfitted with a LiDAR sensor, though it was not used in these experiments. The following three images display surface normals, segmentation results from~\citep{cheng2022masked}, and the planning image, respectively. In the planning image, the left and right safe boundaries are marked to demonstrate the effectiveness of the POVNav framework. The bottom row shows third-person views of the robot at four different timestamps.} \label{Fig:culvert_exp_overview}
\end{figure*}

POVNav facilitates such adaptive navigation by allowing manual selection of navigable classes to achieve desired behaviors. To further explore automation in dynamic navigability, we tested GPT-3.5 as a navigability parser during environment transitions, with surprising results. In \Cref{fig:km_scale_nav}, due to segmentation model inaccuracies, the path was initially labeled as ``sand” or ``earth.” With these classes designated as non-navigable in the urban trail context, the robot was unable to plan a viable path. When provided with these classes, GPT-3.5 suggested a revised navigable set including “path” and “sand” based on context (see timestamps $t1$ and $t2$ in \Cref{fig:km_scale_nav}). At $t3$, when the robot identified “rock” and “stone” classes that also hindered goal alignment, GPT-3.5 adapted the navigable set, considering these classes as traversable in the context of ``urban trails, dense forest paths, open spaces, and roadways.” This adaptive change in visual path planning is evident in \Cref{fig:km_scale_nav}.

\subsection{POVNav for Solving Real-World Problems}
The proposed POVNav framework demonstrates significant potential in addressing practical challenges in real-world applications. This section highlights one such application—automated culvert inspection for infrastructure maintenance—and illustrates how POVNav contributes to its solution. Culvert inspection involves navigating through narrow, often constrained environments where traditional GPS signals are unavailable, and terrain complexity poses considerable challenges. Additionally, the entry size of culverts can restrict the deployment of commonly used robots, necessitating the use of a compact, low-power robot with limited sensing and computational resources. These constraints make POVNav particularly well-suited for this task, as it is computationally efficient and allows for flexible integration with various segmentation techniques according to the available computational resources.

POVNav's vision-based approach supports efficient culvert navigation by utilizing real-time semantic segmentation to identify and follow navigable regions. One of the core advantages of our approach is the ability to redefine “navigability” manually, enabling customization of semantic segmentation models to the unique environment of a culvert. This adaptability ensures that the robot can accurately interpret and traverse safe surfaces. Additionally, POVNav can leverage large language models (LLMs) for enhanced reasoning and contextual adjustments to navigability definitions, enabling improved decision-making in dynamic or poorly defined environments where conventional semantic labels may be insufficient. Furthermore, LLMs can assist in detecting and analyzing structural defects within the culvert.

This integration of semantic segmentation with LLM-enhanced reasoning bolsters the system's robustness, enabling POVNav to perform effectively in challenging real-world scenarios like culvert inspection. As illustrated in \Cref{Fig:culvert_exp_overview}, we deployed POVNav on a robot within a real culvert with a narrow entryway. We evaluated two segmentation techniques—surface normals and Mask2Former~\citep{cheng2022masked}. The Mask2Former method produced incorrect semantic labels within the culvert environment, likely due to the absence of similar scenes in its training data. Nevertheless, by adjusting the navigability criteria, we successfully directed the robot through the culvert. For instance, we defined navigable regions to include classes such as \textit{wall}, \textit{floor}, \textit{person}, \textit{earth}, \textit{ground}, \textit{water}, \textit{rock}, \textit{stone}, \textit{path}, \textit{stairs}, \textit{land}, and \textit{soil}. Interestingly, although the segmentation model incorrectly labeled certain regions as “person,” our adjustable navigability framework allowed POVNav to complete the navigation task effectively by interpreting this label as part of the navigable environment.

\vspace{0.5em}
\noindent
\textbf{Supplementary Video:} A video demonstrating the proposed method in simulation and real-world environments is available at:  
\url{https://youtu.be/wVwIQdXpUM8}

%% file: 6_conclusion.tex
\subsection{Limitations and Future Work}

While POVNav demonstrates promising results in visual semantic navigation, several limitations present opportunities for future enhancement. 

The use of a monocular camera is central to POVNav's lightweight design, making it deployable on low-compute hardware like microcontrollers. However, this configuration lacks the ability to infer depth, restricting environmental understanding to segmented classes without explicit obstacle structure information. Consequently, POVNav heavily relies on the segmentation method's accuracy for obstacle detection and avoidance. We can integrate depth sensors or utilize stereoscopic vision to enable richer environmental context, enhancing both obstacle detection and the robustness of path planning.

When the robot traverses steep inclines or declines, the camera's roll and pitch can alter its field of view, potentially limiting its ability to capture necessary features for feasible path identification and directly affecting the exploration objective of the POVNav. In such cases, the camera may predominantly capture non-navigable surfaces, compromising path planning efficacy. A possible direction to solve this problem is incorporating IMU data to account for roll and pitch effects could maintain a stable navigable viewpoint, helping ensure continuous feature availability in challenging terrains.

While POVNav's paths are efficiently planned in visual space, they may not always be dynamically feasible in real-world scenarios due to the robot's kinematic and dynamic constraints. By learning feasible trajectories in the images space that consider both the robot's dynamics and the physical properties of its surroundings, future iterations of POVNav could yield paths that are better aligned with real-world motion constraints.

Currently, navigability definitions rely on human input or naive GPT-3.5 API utilization, which may not fully capture environmental variability. Segmentation methods are imperfect and can dynamically reclassify the same terrain across different times of day. Our extensive experimentation revealed that navigability definitions shifted within the same environment throughout the day. For instance, in late afternoon, shadows cast by tree branches on the ground were occasionally detected as poles, categorizing them as non-navigable obstacles. This observation highlights the impact of temporal lighting changes on segmentation accuracy. Future advancements could explore the integration of physical properties, such as IMU-based vibration analysis over varied surfaces, to assess factors like friction and ground consistency. These attributes could provide additional context, helping the system adaptively refine traversability definitions and thereby improve navigation robustness.

These enhancements present a path forward for refining POVNav into a more versatile and resilient navigation framework, particularly suited to environments with dynamic conditions and constrained compute resources.

Finally, while the current control law ensures bounded velocities, it does not explicitly enforce dynamic feasibility under traction-limited conditions. In high-speed navigation scenarios, especially on low-friction or uneven surfaces, the combined lateral and longitudinal acceleration can exceed the platform’s traction capacity, resulting in loss of control. As a future direction, we plan to incorporate a traction-aware control mechanism into POVNav. This would involve adaptively modulating the translational velocity $v$ based on the commanded angular velocity $\omega$ and estimated friction limits, thereby enforcing the dynamics constraint during motion planning and control. This extension will allow POVNav to operate safely and robustly at higher speeds and in more challenging terrain, improving real-world deployment capabilities.

\section{Conclusion}
This paper presents a visual semantic navigation system, POVNav, designed to empower autonomous robots with robust navigation capabilities in diverse real-world environments. By leveraging semantic segmentation to distinguish traversable regions from obstacles, POVNav enables robots to achieve near-human-like navigation behavior, essential for complex and dynamic applications. The framework's modular design integrates the flexibility of end-to-end image-to-action mapping without the need for specialized or semantic maps, making it both adaptable and efficient. It further allows for the integration of advanced learning-based semantic understanding techniques while introducing mechanisms to mitigate errors when confronted with out-of-distribution data—a common occurrence in real-world deployment. Additionally, while POVNav is compatible with sophisticated segmentation techniques, it retains the flexibility to operate with simpler segmentation methods, such as color-based or surface-normal approaches, underscoring its broad applicability. 
Also, each component of POVNav can be further refined or replaced to enhance its suitability for increasingly challenging scenarios, underscoring its versatility and practical value.

We conducted extensive simulation and field experiments to validate each component of POVNav. In simulations, the proposed system demonstrated superior performance compared to existing methods, excelling in success rate, path length efficiency, computational speed, and resilience to dynamic obstacles. Field experiments were performed in varied settings—indoor navigation with a Jackal robot utilizing noisy surface-normal segmentation, which tested POVNav's capacity for waypoint navigation and obstacle avoidance under a highly unstable visual horizon. Outdoor trials, conducted with both Husky and Unitree Go1 robots, included navigating harsh weather conditions with minimal drift or instability, despite the extreme cold affecting the goal direction sensors. These experiments employed basic color-based segmentation to navigate stark black-and-white visual environments, illustrating the method's adaptability.

We performed long-range navigation in urban trails, sparse and dense forest trails, structured roads, and populated campus pathways. These tests were conducted across different times of day, seasons (fall and winter), and varying environmental conditions, revealing the dynamic nature of current state-of-the-art segmentation models. Our work highlights the potential of vision-language models to dynamically adjust navigability definitions in response to shifting environmental semantics, a capability that bridges the limitations of conventional segmentation modules. Frame rates for segmentation during these experiments varied between 1 Hz and 30 Hz, with POVNav successfully adapting to these constraints. 
The reliability demonstrated across more than 100 successful real-world navigation trials, spanning different robots, environments, and segmentation methods motivated us to extend this approach to real-world applications such as culvert inspection.

\section{Acknowledgement}

The authors extend their appreciation to Md. Al-Masrur Khan and Mahmoud Ali for their assistance in conducting the hardware experiment in the wild. 
Additionally, the authors sincerely thank all anonymous reviewers for their invaluable and constructive feedback, which has significantly contributed to the improvement and enhanced quality of this manuscript.

\section{Declaration of Conflicting Interests}
The authors declared no potential conflicts of interest with
respect to the research, authorship, and/or publication of this
article. 

\section{Funding}
The authors disclosed receipt of the following financial
support for the research, authorship, and/or publication of this
article. 
This work was partially supported by  
(1)~NSF: CAREER: Autonomous Live Sketching of Dynamic Environments by Exploiting Spatiotemporal Variations (grant no 2047169); 
(2) U.S. Army Research Office and was accomplished under Cooperative Agreement
Number W911NF-22-2-0018: Scalable, Adaptive, and Resilient Autonomy (SARA);
(3) U.S. Army Engineer Research and Development Center (grant no W912HZ2320003, subaward no M2303647). 
The views and conclusions contained in this document are those of
the authors and should not be interpreted as representing the
official policies, either expressed or implied, of the U.S. Government. The U.S. Government is authorized to reproduce and distribute reprints for Government purposes notwithstanding any copyright notation
herein. 